\documentclass{article}

     \PassOptionsToPackage{numbers, compress}{natbib}

\usepackage[final]{neurips_2025}

\usepackage[utf8]{inputenc} 
\usepackage[T1]{fontenc}    
\usepackage{hyperref}       
\usepackage{url}            
\usepackage{booktabs}       
\usepackage{amsfonts}       
\usepackage{nicefrac}       
\usepackage{microtype}      
\usepackage{xcolor}         

\usepackage{amsmath}
\usepackage{diagbox}
\usepackage{caption}
\usepackage{amsfonts,amssymb}
\usepackage{subfigure}
\usepackage{graphicx}
\usepackage{bm}
\usepackage{multirow}
\usepackage[ruled,lined,commentsnumbered]{algorithm2e}
\usepackage{hyperref}
\usepackage{booktabs}
\usepackage{bbm}
\usepackage[inkscapelatex=false]{svg}
\usepackage{pifont}
\usepackage{stfloats}
\usepackage{multicol}
\usepackage{float}
\usepackage{array}
\newcommand{\PreserveBackslash}[1]{\let\temp=\\#1\let\\=\temp}
\newcolumntype{C}[1]{>{\PreserveBackslash\centering}p{#1}}
\newcolumntype{R}[1]{>{\PreserveBackslash\raggedleft}p{#1}}
\newcolumntype{L}[1]{>{\PreserveBackslash\raggedright}p{#1}}

\usepackage{mathrsfs}
\usepackage{wrapfig}

\usepackage{booktabs}   
\usepackage{caption}    
\usepackage{subcaption} 

\title{FastLongSpeech: Enhancing Large Speech-Language Models for Efficient Long-Speech Processing}

\author{
  Shoutao Guo$^{1,3}$, Shaolei Zhang$^{1,3}$, Qingkai Fang$^{1,3}$, Zhengrui Ma$^{1,3}$, \\ \textbf{Min Zhang}$^{4}$, \textbf{Yang Feng}$^{1,2,3}$\thanks{ Corresponding author: Yang Feng.} \\
        \textsuperscript{\rm 1}{Key Laboratory of Intelligent Information Processing,} \\ Institute of Computing Technology, Chinese Academy of Sciences (ICT/CAS) \\
    { \textsuperscript{\rm 2} {Key Laboratory of AI Safety, Chinese Academy of Sciences}} \\
    { \textsuperscript{\rm 3} {University of Chinese Academy of Sciences, Beijing, China}} \\
{ \textsuperscript{\rm 4} {School of Future Science and Engineering, Soochow University}} \\
  {$\;\:$\texttt{\href{mailto:guoshoutao22z@ict.ac.cn}{guoshoutao22z@ict.ac.cn},\href{mailto:zhangshaolei20z@ict.ac.cn}{zhangshaolei20z@ict.ac.cn},\href{mailto:fengyang@ict.ac.cn}{fengyang@ict.ac.cn}}}
}

\begin{document}

\maketitle

\begin{abstract}
The rapid advancement of Large Language Models (LLMs) has spurred significant progress in Large Speech-Language Models (LSLMs), enhancing their capabilities in both speech understanding and generation. While existing LSLMs often concentrate on augmenting speech generation or tackling a diverse array of short-speech tasks, the efficient processing of long-form speech remains a critical yet underexplored challenge. This gap is primarily attributed to the scarcity of long-speech training datasets and the high computational costs associated with long sequences.
To address these limitations, we introduce FastLongSpeech, a novel framework designed to extend LSLM capabilities for efficient long-speech processing without necessitating dedicated long-speech training data. FastLongSpeech incorporates an iterative fusion strategy that can compress excessively long-speech sequences into manageable lengths. To adapt LSLMs for long-speech inputs, it introduces a dynamic compression training approach, which exposes the model to short-speech sequences at varying compression ratios, thereby transferring the capabilities of LSLMs to long-speech tasks.
To assess the long-speech capabilities of LSLMs, we develop a long-speech understanding benchmark called LongSpeech-Eval. Experiments show that our method exhibits strong performance in both long-speech and short-speech tasks, while greatly improving inference efficiency \footnote{The Code is at \url{https://github.com/ictnlp/FastLongSpeech.git}.}.
\end{abstract}

\section{Introduction}

Benefiting from the advancement of Large Language Models (LLMs) \citep{openai2024gpt4ocard, grattafiori2024llama3herdmodels, deepseekai2024deepseekv3technicalreport}, Large Speech-Language Models (LSLMs) have also made significant strides by extending the speech capabilities of LLMs. By harnessing the knowledge and reasoning abilities of LLMs, LSLMs can directly comprehend speech signals, perform analysis and reasoning, and achieve superior performance in a diverse of tasks such as speech recognition, speech translation, and speech understanding \citep{tang2024salmonn, chu2024qwen2audiotechnicalreport, chen2025minmomultimodallargelanguage}. The ability to process and understand diverse speech signals has emerged as a key research focus in LSLMs \citep{chu2023qwenaudioadvancinguniversalaudio}.

To handle speech inputs, traditional methods \citep{shen2023hugginggptsolvingaitasks, huang2023audiogptunderstandinggeneratingspeech} typically employ a cascaded pipeline, where speech is first transcribed into text and then processed by LLMs. However, these approaches suffer from error propagation and discard valuable paralinguistic information \citep{wang2024blspbootstrappinglanguagespeechpretraining}. To overcome these drawbacks, recent research \citep{zhang2023speechgptempoweringlargelanguage, microsoft2025phi4minitechnicalreportcompact,kimiteam2025kimiaudiotechnicalreport} has shifted towards an end-to-end paradigm, enabling LSLMs to directly process and reason with speech signals. These methods can be broadly divided into two categories. On one hand, some approaches \citep{tang2024salmonn, fang2024llamaomniseamlessspeechinteraction, xu2025qwen25omnitechnicalreport, fang2025llamaomni2llmbasedrealtimespoken} such as Qwen2-Audio \citep{chu2024qwen2audiotechnicalreport} align the output spaces of pre-trained audio encoders with the embedding of LLMs, which allows speech inputs to be accommodated while transferring partial capabilities of the employed LLMs. At the same time, other methods \citep{zhang2023speechgptempoweringlargelanguage, zeng2024glm4voiceintelligenthumanlikeendtoend} involves discretizing
the speech to discrete units, which allows LSLMs to handle speech units similar to text tokens. Despite these advancements, current LSLMs are largely constrained to processing short speech segments, typically under 30 seconds \citep{chu2024qwen2audiotechnicalreport, fang2025llamaomniseamlessspeechinteraction}. Only a few LSLMs \citep{microsoft2025phi4minitechnicalreportcompact} have achieved a processing duration of 30 minutes on speech summarization tasks by relying on the construction of extensive, specialized training datasets.

The processing of long-form speech by LSLMs remains a largely unexplored area, primarily due to two significant challenges. First, unlike the abundance of diverse short-speech datasets \citep{chung18b_interspeech, wang21s_interspeech, 2023jointaudiospeechunderstanding}, there is a scarcity of training data specifically for long-speech alignment and instruction, and the generation of such data is costly. Second, long-speech sequences impose substantial computational demands on LSLMs. The sequence of speech representations is often more than four times longer than its text equivalents for the same content \citep{zeng2024glm4voiceintelligenthumanlikeendtoend}. which, in the context of long-form speech, leads to significantly higher computational costs. Therefore, LSLMs face considerable challenges in modeling long-speech sequences, stemming from both a scarcity of training data and increased computational costs. These challenges limit the exploration and application of LSLMs in long-speech processing.


To address the above challenges, we propose FastLongSpeech, a novel framework designed to extend the capabilities of LSLMs to long-speech processing, leveraging only short-speech training data. We utilize Qwen2-Audio\footnote{For convenience, Qwen2-Audio refers to Qwen2-Audio-7B-Instruct throughout this paper.} \citep{chu2024qwen2audiotechnicalreport}, the currently representative LSLM, as our foundational speech-language model. To enable long-speech processing, FastLongSpeech incorporates a speech extractor module on top of the audio encoder \citep{radford2022robustspeechrecognitionlargescale}. This module employs our proposed iterative fusion strategy to compress the speech representations output by the audio encoder, preserving essential temporal information while reducing redundancy. The resulting condensed speech representations are then used by LLM for comprehension and reasoning. In our method, the speech extractor significantly reduces the sequence length of the condensed representations, thereby lowering the computational costs for subsequent LLM.

To further adapt LSLM for long-speech processing, FastLongSpeech employs a two-stage training approach. In the first stage, a CTC loss \citep{graves2006connectionist} is introduced within the extractor module to measure the text density of the input speech representations, which is utilized for the iterative fusion strategy. The second stage introduces a dynamic compression training method, enabling LLM to effectively adapt to condensed speech representations. This stage leverages existing short-speech data and dynamically adjusts the compression ratios in the iterative fusion strategy to transfer the understanding and reasoning capabilities of LSLMs to long-speech tasks. 

To evaluate the long-speech understanding capabilities of LSLMs, we also develop a benchmark, called LongSpeech-Eval. Experiments show that FastLongSpeech achieves efficient speech processing on both long-speech and short-speech benchmarks, and can balance efficiency and effectiveness to meet different requirements.

\section{Background}
Our method builds upon Qwen2-Audio \citep{chu2024qwen2audiotechnicalreport} and incorporates CTC loss \citep{graves2006connectionist}. We provide a brief overview of these key components below.

\paragraph{Qwen2-Audio} 
Qwen2-Audio is a large-scale audio-language model capable of comprehending various audio signals, performing reasoning, and generating text responses. It consists of three modules: an audio encoder \citep{radford2022robustspeechrecognitionlargescale}, an LLM \citep{bai2023qwentechnicalreport}, and an adaptor. The adaptor serves to align the output of the audio encoder with the embeddings of LLM, enabling the LLM to process speech inputs. Following extensive pre-training, supervised fine-tuning, and DPO \citep{rafailov2023direct}, Qwen2-Audio demonstrates remarkable performance across a wide range of speech tasks. Given a raw waveform $\mathbf{s}$ = $(s_1, ..., s_N)$ sampled at 16 kHz, the audio encoder produces a sequence of speech representations $\mathbf{h}$ = $(h_1, ..., h_{J})$ with 25 Hz frame rate.
The LLMs then generate the text response $\mathbf{y}$ based on $\mathbf{h}$ and an instruction $\mathbf{x}$:
\begin{equation}
\label{infer}
    p(\mathbf{y}|\mathbf{x}, \mathbf{h}) = \sum\limits_{i=1}^{I} p(y_i \mid \mathbf{y}_{<i}, \mathbf{x}, \mathbf{h}),
\end{equation}
where $p(y_i \mid \mathbf{y}_{<i}, \mathbf{x}, \mathbf{h})$ denotes probability distribution of the next token.

\paragraph{CTC} 
Connectionist Temporal Classification (CTC) \citep{graves2006connectionist} is widely used in Automatic Speech Recognition (ASR), where the input is typically longer than the corresponding output \citep{guo2023simultaneousmachinetranslationtailored}. To align speech with the transcript, CTC introduces a blank token into the vocabulary and defines the possible output as an alignment $\mathbf{a}$. Each time step in this alignment corresponds to either a blank token or a non-blank token. The alignment has the same length as the speech sequence and can be reduced to the final transcript through a collapse function $\Gamma^{-1}$. During training, CTC employs an efficient dynamic programming algorithm to maximize the probability of all possible alignments corresponding to the ground-truth transcript $\mathbf{c}$:
\begin{equation}
\label{ctc}
\mathcal{L}_{ctc} = - \log \sum\limits_{\mathbf{a} \in \Gamma(\mathbf{c})} p_{ctc}\!\left ( \mathbf{a}\mid \mathbf{h} \right ),
\end{equation}
where $\Gamma(\mathbf{c})$ denotes the set of all alignments corresponding to $\mathbf{c}$, and $\mathbf{h}$ denotes the sequence of speech representations produced by audio encoder \citep{radford2022robustspeechrecognitionlargescale}. In this paper, we leverage the output distribution of CTC to quantify the text density of speech representations $\mathbf{h}$, which is subsequently utilized in the iterative fusion strategy.

\section{Method}
In this section, we introduce the architecture of FastLongSpeech, with a particular emphasis on its novel speech extractor module. To empower LSLM with the ability to learn speech compression techniques and adapt to long-speech processing, we present an innovative two-stage training approach. Additionally, we introduce the LongSpeech-Eval benchmark, designed to evaluate the long-speech understanding capabilities of LSLMs.

\begin{figure*}[t]
    \centering
    \includegraphics[width=\textwidth]{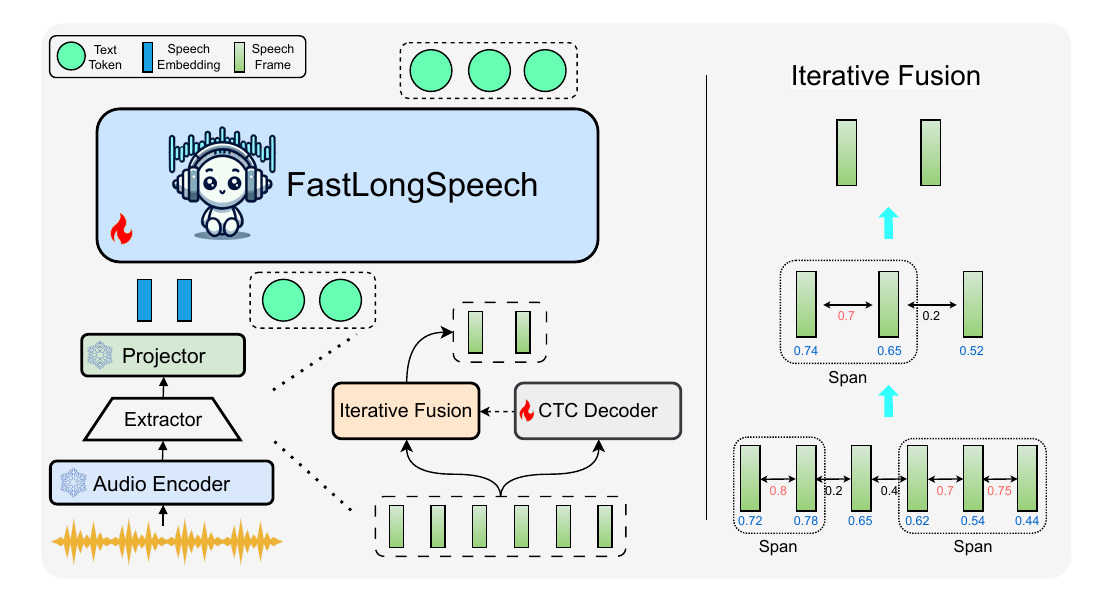}
    \caption{Architecture of FastLongSpeech. The left panel illustrates that FastLongSpeech generates a response based on the input speech and text instruction. The right panel details the iterative fusion strategy, where numbers between adjacent frames denote similarity scores and numbers below frames represent content density.}
    \label{fig-model} 
\end{figure*}

\subsection{Model Architecture}
Figure \ref{fig-model} illustrates the model framework of FastLongSpeech. Building upon Qwen2-Audio, we incorporate an advanced extractor module, which features a CTC \citep{graves2006connectionist} decoder and employs our proposed iterative fusion strategy to condense speech representations. The workflow of FastLongSpeech proceeds as follows. Given a waveform $\mathbf{s}$ sampled at 16 kHz, the audio encoder converts it into a mel-spectrogram and processes it through convolution and transformer layers \citep{radford2022robustspeechrecognitionlargescale}, yielding a sequence of speech representations $\mathbf{h}$ with 25 Hz frame rate. Subsequently, the extractor module processes $\mathbf{h}$ using an iteration fusion strategy to produce the condensed representations $\mathbf{h^{\prime}}$, whose length is within the speech window of LLM. The speech window denotes the maximum length of the speech representations $\mathbf{h}$ during training, specifically the maximum number of speech frames. LLM then utilizes the condensed representations $\mathbf{h^{\prime}}$ along with the instruction $\mathbf{x}$ to generate the response $\mathbf{y}$, as described in Eq.(\ref{infer}).

\subsection{Iterative Fusion}
To facilitate efficient long-speech processing, we introduce an extractor to compress lengthy and sparse speech representations \citep{zeng2024glm4voiceintelligenthumanlikeendtoend} into more compact forms. Our primary objective is to minimize information loss during compression, preserving essential temporal information while reducing redundancy. Thus, our method needs to retain speech representations that contain more textual content, while discarding excessively similar adjacent speech frames. To achieve this, we propose an iterative fusion strategy that incrementally merges selected representations based on content density and similarity between adjacent frames, ultimately yielding condensed representations.

We first define the metrics to measure content density and frame similarity, followed by an introduction of our iterative fusion strategy. For a given speech frame $h_j$, content density is generally associated with the amount of textual information it contains \citep{ren-etal-2020-simulspeech, zhang-feng-2023-end}. To quantify this, we employ a CTC decoder, whose output distribution provides the probabilities of the speech frame being classified as either a blank or a non-blank token. Consequently, the content density of $h_j$ is derived from the sum of probabilities for non-blank tokens:
\begin{equation}
\label{weight}
    d_j = \sum_{a_j \neq \epsilon} p_{ctc}(a_j \mid h_j),
\end{equation}
where $\epsilon$ denotes the blank token. Frame similarity, on the other hand, captures the overlap in content between adjacent speech frames. We measure this using cosine similarity between $h_j$ and $h_{j+1}$:
\begin{equation}
\label{similar}
    e_{j,j+1} = \frac{h_j h_{j+1}}{\lvert h_j \rvert \lvert h_{j+1} \rvert}.
\end{equation}

After introducing these measurements, we provide an overview of our iterative fusion strategy in Figure \ref{fig-model}. For $m$-th iteration, we first determine the length $T(m)$ of current speech representations and the length for the next iteration:
\begin{equation}
    T(m\!+\!1) =
\begin{cases} 
    \lfloor T(m) / 2 \rfloor, & \text{if } T(m) > 2L \\
    L, & \text{if } T(m) \leq 2L 
\end{cases},
\end{equation}
where $L$ denotes the final target length of condensed representations $\mathbf{h}^{\prime}$. The number of speech frames to be reduced is then calculated as:
\begin{equation}
    r(m) = T(m) - T(m\!+\!1).
\end{equation}
Subsequently, we utilize the similarity metric between adjacent frames, as defined in Eq.(\ref{similar}) to identify the $r(m)$ most similar pairs of adjacent frames. The consecutive identified frames are grouped into a span, as illustrated in Figure \ref{fig-model}. For each span, we employ a weighted fusion approach, leveraging the content density in Eq.(\ref{weight}) as weights to merge all frames within the span into a single compressed speech frame. This process yields a new sequence of speech representations with length $T(m\!+\!1)$, comprising both the newly condensed frames and the remaining uncompressed frames.
If $T(m\!+\!1)$ still exceeds the target length $L$, we initiate another iteration. Otherwise, the resulting speech representations from this round constitute the final condensed representations $\mathbf{h}^{\prime}$. 

This iterative fusion strategy effectively reduces the length of speech representations by half in each iteration, facilitating efficient speech processing.





\subsection{Training Method}
After introducing the iterative fusion strategy, we further present a two-stage training method to enable FastLongSpeech to perform long-speech tasks. To facilitate the generation of condensed representations through the iterative fusion strategy, the first stage of training focuses on the ASR task, allowing FastLongSpeech to learn the measurement of content density. Building on this, the second stage incorporates our proposed dynamic compression training method, which helps LLMs adapt to short-speech condensed representations with varying compression ratios, thereby transferring short-speech capabilities to long-speech processing.

\paragraph{CTC Training} In the first stage, we aim to leverage the ASR task to enable FastLongSpeech to recognize the amount of textual information in speech representations, namely content density. To achieve this, we introduce a CTC decoder in the extractor module, which is trained using the CTC loss \citep{graves2006connectionist} as shown in Eq.(\ref{ctc}). This allows us to utilize the generation distribution of the CTC decoder to measure the content density of speech representations. In this stage, we only train the CTC decoder.

\paragraph{Dynamic Compression Training} In the second stage, we introduce a novel dynamic compression training method. The introduction of this method is based on two considerations. First, it enables the LLM to adapt to condensed representations $\mathbf{h}^{\prime}$ with different compression ratios. Second, the current <$\mathbf{s}$, $\mathbf{x}$, $\mathbf{y}$> triplet training data primarily contains short-speech clips, which are typically under 30 seconds in duration \citep{lin2024speechprunecontextawaretokenpruning}. By sampling the length $L$ of condensed representations \citep{Guo_2024}, LSLM can maintain its perception of speech sequences corresponding to the length of its speech window, without avoiding excessive bias towards overly condensed speech sequences. The dynamic compression training method is as follows:
\begin{equation}
    L_{dct} =  \!-\! \sum\limits_{L \sim \mathcal{U}(\emph{\rm{L}})} \log p(\mathbf{y} \mid \mathbf{x}, \mathrm{IF}(\mathbf{h}, L)),
\end{equation}
where $L$ is uniformly sampled from $\emph{\rm{L}}$, which is a set of hyperparameters. $\mathrm{IF}(\mathbf{h}, L)$ represents applying the iterative fusion operation on the speech representations $\mathbf{h}$ to obtain the condensed representations of length $L$. After the training process, FastLongSpeech can transfer the short-speech capabilities of LSLMs to long-speech tasks.

\subsection{LongSpeech-Eval Benchmark}
After introducing FastLongSpeech, we aim to evaluate its capacity to handle long-speech inputs. Due to the lack of benchmarks for assessing the long-speech capabilities of LSLMs, we construct LongSpeech-Eval. This benchmark is based on the MultiFieldQA-En and NarrativeQA subsets of LongBench \citep{bai-etal-2024-longbench}, a comprehensive long context understanding benchmark. This benchmark assesses the ability of LLM to answer questions based on the long document.

To construct LongSpeech-Eval, we aim to convert the document into speech. We first employ Llama3.1-70B-Instruct\footnote{\url{https://huggingface.co/meta-llama/Llama-3.1-70B-Instruct}} to filter out samples containing numerous formulas or non-English characters. Subsequently, GPT-4o \citep{gpt-4o} is utilized to summarize and polish the document into a spoken format. Llama3.1-70B-Instruct is then used to answer questions based on the spoken-form document, with inappropriate samples manually discarded. For the remaining samples, we then synthesize speech for the spoken-form document using Orca\footnote{\url{https://github.com/Picovoice/orca.git}}. Consequently, each sample in the LongSpeech-Eval consists of the synthesized speech along with the corresponding questions and answers. More details are provided in \textbf{Appendix \ref{longbench}}.

\section{Experiments}
\subsection{Datasets}

For the training data in the first stage, we utilize the ASR data, which contain 960 hours of LibriSpeech \citep{7178964} data and 3k hours of data sampled from MLS \citep{Pratap_2020}.
In the second training stage, our training data primarily originates from three datasets following the Spoken QA format: \textbf{OpenASQA} \citep{10389742}, \textbf{LibriSQA} \citep{zhao2024librisqanoveldatasetframework}, and \textbf{Common Voice} \citep{ardila-etal-2020-common}. All employed samples used for training are under 30 seconds in duration. For OpenASQA, we use the Open-Ended Speech AQA subset (5.9k hours), which covers diverse question types including content, speaker style, and emotion. For LibriSQA, we incorporate the 360-hour training set. For Common Voice, we adapt the English subset (1.7k hours) to the spoken QA format by generating transcription instructions via ChatGPT and using the original transcripts as the ground-truth answers.

For evaluation, we employ a diverse set of nine datasets spanning five distinct tasks to comprehensively assess performance:

\textbf{Short-Speech Spoken QA}: We utilize three datasets: the speech\_QA\_iemocap (AIR-Bench) \citep{yang-etal-2024-air}, the LibriSQA test set \citep{zhao2024librisqanoveldatasetframework}, and the LibriTTS test subset from OpenASQA \citep{10389742}. The three datasets contain rich set of QA pairs involving paralinguistic information. We utilize this task to evaluate the effectiveness of various speech fusion methods under different compression ratios.

\textbf{Long-Speech Spoken QA}: We leverage our proposed LongSpeech-Eval benchmark to assess the performance of different methods in long-speech understanding scenarios.

\textbf{Spoken Dialogue Understanding}: We evaluate the inference efficiency of our method using speech\_dialogue\_QA\_fisher subset from AIR-Bench.

\textbf{Emotion Recognition}: We leverage the MELD dataset \citep{poria-etal-2019-meld} to benchmark our method against other efficiency method \citep{lu2024fastadaspmultitaskadaptedefficientinference} under diverse efficiency scenarios.

\textbf{ASR}: We use the LibriSpeech \citep{7178964} test-clean and test-other, and GigaSpeech \citep{chen2021gigaspeechevolvingmultidomainasr} test set to evaluate ASR performance. The ASR task utilizes short-speech samples, which evaluate the capacity to extract complete content from the speech.

\textbf{Speech Information Retrieval}: We introduce SPIRAL-H benchmark, which is introduced in the paper of SpeechPrune \citep{lin2025speechprunecontextawaretokenpruning} for long speech information retrieval.

More details of the training and evaluation dataset are in \textbf{Appendix \ref{dataset}}.


\subsection{System Settings}
Given the current scarcity of long-speech LSLMs, we implement several baseline approaches in addition to our FastLongSpeech. 

\textbf{Random} method randomly selects frames from speech representations and arranges them sequentially to serve as conditioning input for LSLM.

\textbf{AvgPool} method sequentially selects a fixed number of speech frames to form segments and then apply average operation within each segment.

\textbf{MostSim} method first selects the most similar consecutive speech frames and then applies the average pooling operation to each set of adjacent representations.

\textbf{NTK-RoPE} method modifies the base Rotary Position Embedding (RoPE) \citep{su2023roformerenhancedtransformerrotary} of LLMs to an NTK-Aware Scaled RoPE \citep{bloc97_2023}, thereby extending the speech window of Qwen2-Audio to match the context length of its LLM. This method is exclusively utilized for long-speech reasoning tasks.

\textbf{Cascaded} first uses Whisper-Large-V3 \citep{radford2022robustspeechrecognitionlargescale} to transcribe the audio, and then pass the resulting context text to Qwen-7B-Chat (since Qwen2-Audio-7B is based on Qwen-7B-Chat).

\textbf{Baseline} refers to the direct application of vanilla Qwen2-Audio for inference, which is not employed in long-speech inference tasks. 

\textbf{FastLongSpeech} refers to our proposed method.

We use Qwen2-Audio-7B-Instruct\footnote{\url{https://huggingface.co/Qwen/Qwen2-Audio-7B-Instruct}} as the base LSLM for all methods, with the length of speech window as 750. Besides, we also extend our method to vanilla Qwen2.5-Omni \citep{xu2025qwen25omnitechnicalreport} without dynamic compression training to verify the effectiveness of our method. In the first stage of training for our FastLongSpeech, we only train the CTC decoder. In the second stage, we experimentally set $\mathbf{L}$ to \{750, 400, 200, 100, 50, 25, 12\} and fine-tune the LLM of Qwen2-Audio using LoRA \citep{hu2021lora}. For a fair comparison, we also fine-tune Qwen2-Audio for all methods except the Baseline and FastLongSpeech, using the same training data and implementation settings as used for FastLongSpeech. All methods employ the original prompt template from Qwen2-Audio. For more training details, please refer to the \textbf{Appendix \ref{experiment}}.

\begin{figure*}[t]
\centering
\subfigure[speech\_QA\_iemocap]
{
\label{main_1}
\includegraphics[width=1.75in]{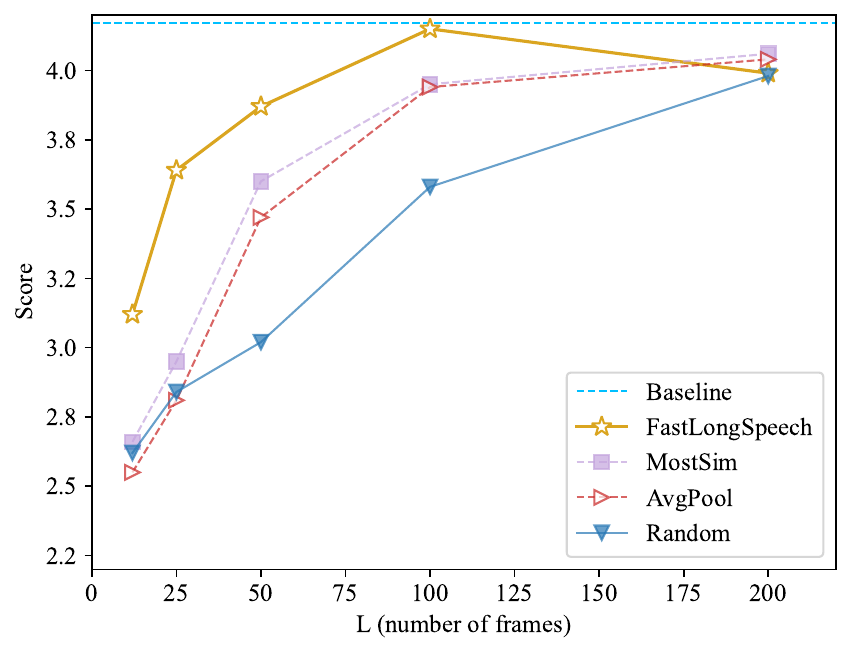}
}\hspace{-0.2cm}
\subfigure[LibriTTS (OpenASQA)]{
\label{main_2}
\includegraphics[width=1.75in]{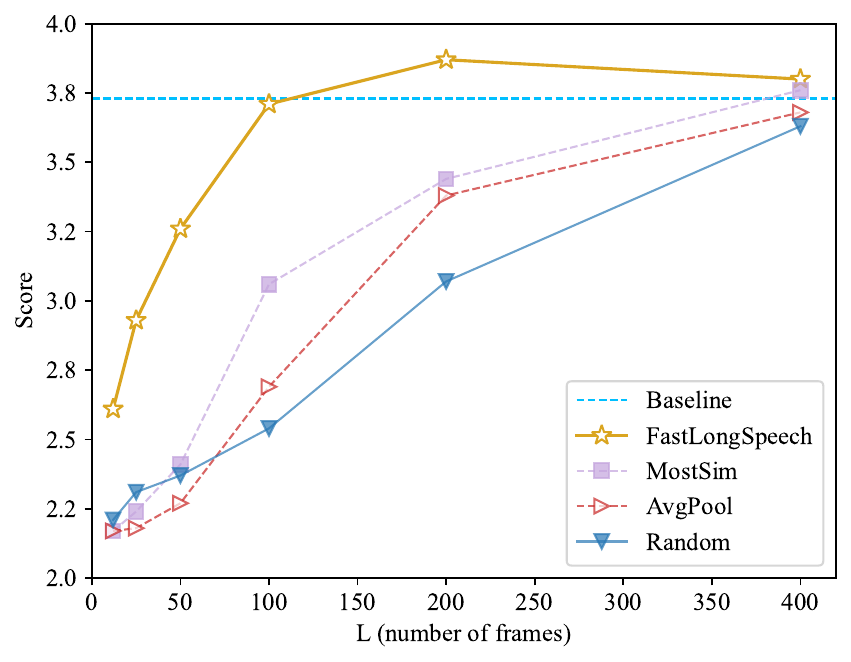}
}\hspace{-0.2cm}
\subfigure[LibriSQA]{
\label{main_3}
\includegraphics[width=1.75in]{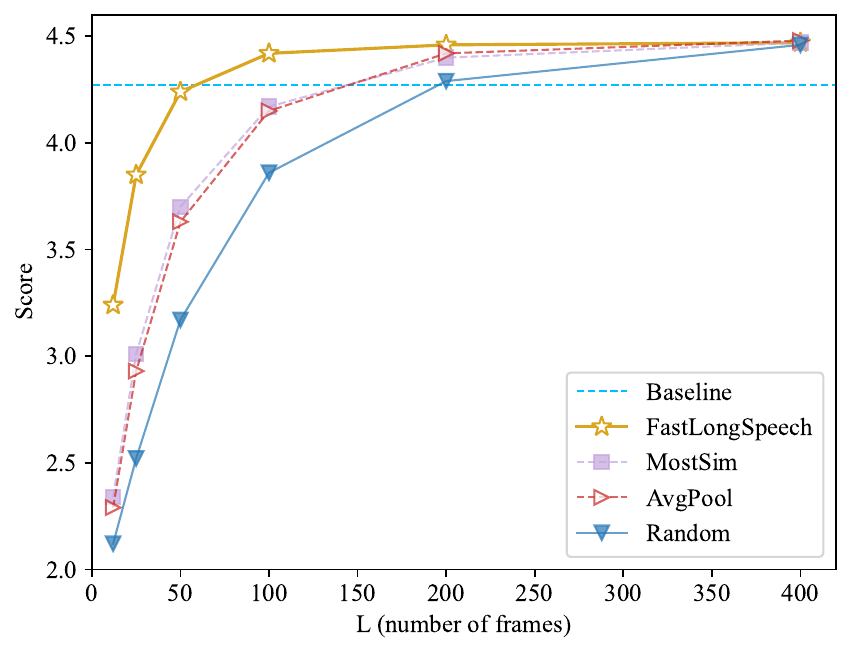}
}

\caption{Performance of diverse speech fusion methods in the short-speech spoken QA tasks. The score is derived from the LLM evaluating the quality of responses based on the questions and ground-truth answers.
The baseline model utilizes a speech window of 750 frames. For the methods other than Baseline, we regulate the compression ratio by adjusting the target length $L$ of the condensed speech representations. In the experiments, a smaller value of $L$ corresponds to a higher compression ratio. A higher score indicates a better quality of the responses.}
\label{main_res}
\end{figure*}















During evaluation, we use greedy search for all methods and control compression ratios by adjusting the target length $L$. For long-speech inference, we split the input speech into a series of 30-second clips, which are processed by the audio encoder and then combined into a complete sequence of speech representations in temporal order. To evaluate the performance, we employ various metrics tailored to each task. For the Spoken QA and Spoken Dialogue Understanding task, we use Llama3.1-70B-Instruct \citep{grattafiori2024llama3herdmodels} to score responses on a scale of 1 to 5, with the scoring template available in the \textbf{Appendix \ref{evaluation_template}}. For the ASR task, we use Word Error Rate (WER) to assess the accuracy of the generated transcripts. For Emotion Recognition task, we use the Accuracy (ACC) metric to evaluate the performance.


\subsection{Main Results}
We evaluate the performance of our method on short-speech and long-speech spoken QA tasks.

\begin{wraptable}{r}{4.6cm}
\vspace{-0.08in} 
\caption{Performance of various speech methods on long-speech spoken QA task.}






\label{data_type}
\small
\centering
\label{long_speech}
\begin{tabular}{cccc}
\toprule[1.2pt]
\textbf{Method} & \textbf{Score ($\uparrow$)}   \\ 
\cmidrule(lr){1-1} \cmidrule(lr){2-2}

Random & 2.54 \\

Similar & 3.08 \\

AvgPool & 3.10 \\

NTK-RoPE & 3.44 \\

\textbf{FastLongSpeech} & \textbf{3.55} \\

 \bottomrule[1pt]   
\end{tabular}
\vspace{0in} 
\end{wraptable}
For the short-speech spoken QA task, Figure \ref{main_res} illustrates the performance of various methods across three datasets. Our FastLongSpeech method consistently outperforms other methods on all three datasets under different speech compression ratios, maintaining high response quality even at a 30-fold compression ratio ($L$ = 25). Unlike the Random method, which arbitrarily discards speech representations, other methods consider all temporal information when compressing speech representation, resulting in improved generation quality \citep{lin2024speechprunecontextawaretokenpruning}. Compared to AvgPool and MostSim methods, our method more effectively eliminates redundant information while preserving highly informative speech representations, leading to better performance across various compression ratios. 
Notably, when $L$ equals 12, other speech fusion methods exhibit similarly suboptimal performance, while our method maintains a substantial performance advantage. We attribute this superiority to our novel iterative fusion strategy and dynamic compression training approach. 
Furthermore, compared to vanilla Qwen2-Audio, our method achieves comparable performance with a shorter sequence of speech representations, demonstrating higher efficiency.

For the long-speech spoken QA task, our method outperforms other approaches in generation quality, as evidenced in Table \ref{long_speech}. To handle the long-speech input, methods such as Random, Similar, and AvgPool employ their respective fusion techniques to compress the speech representations within the speech window. However, these approaches yield suboptimal generation quality, primarily due to ineffective fusion strategies and misaligned training methods. In contrast, NTK-RoPE expands the speech window of LSLM to the context length of LLM, thereby preserving more speech information and achieving improved performance. Furthermore, our method leverages a more effective speech fusion strategy coupled with a dynamic compression training approach, transferring the short-speech reasoning capabilities of LSLMs to the long-speech domain. Notably, despite utilizing the same speech window size as Qwen2-Audio \citep{chu2024qwen2audiotechnicalreport}, our method achieves optimal performance in long-speech comprehension tasks with greater efficiency than NTK-RoPE.

\section{Analysis}
To provide a comprehensive evaluation of our approach, we conduct extensive analyses. We then introduce each analytical experiment in detail.

\begin{wraptable}{r}{5.9cm}
\vspace{-0.1in} 
\caption{The ablation experiments of our method on long-speech benchmark. ``w/o DCT'’ replaces Dynamic Compression Training method with standard fine-tuning approach. ``w/o Iterative Fusion'’ eliminates the multiple iterations in the iterative fusion. ``w/o Content Density'’ substitutes the method of merging all speech frames within the same span with an average pooling operation.}
\small
\centering
\label{ablation_study}
\begin{tabular}{l c  c} \toprule[1.2pt]
\textbf{$\;\;\;\;\;\;\;$Method} & \textbf{Score ($\uparrow$)}  \\ 

\cmidrule(lr){1-1} \cmidrule(lr){2-2}

\textbf{FastLongSpeech} & \textbf{3.55} \\


$\;\;$ w/o DCT & 3.33 \\

$\;\;$ w/o Iterate Fusion &  3.41 \\

$\;\;$ w/o Content Density &3.28 \\

 \bottomrule[1pt]
\end{tabular}
\vspace{-0.1in} 
\end{wraptable}

\subsection{Ablation Study}
To gain a comprehensive understanding of the contributions made by different components in our approach, we conduct detailed ablation experiments. As shown in Table \ref{ablation_study}, both the iterative fusion and dynamic compression training strategies proposed in FastLongSpeech significantly enhance the performance of LSLMs on long-speech reasoning tasks.
First, the dynamic compression training strategy effectively transfers the short-speech capabilities of LSLMs to long-speech scenarios, utilizing only short-speech data. This approach enables LLMs to adapt to condensed representations at varying compression ratios and mitigate over-reliance on excessively compressed speech representations. Consequently, FastLongSpeech can compress long-speech representations to fit within the speech window length, facilitating efficient long-speech processing at high compression ratios.
Moreover, multiple iterations in the iterative fusion approach lead to substantial improvements in generation quality. This finding underscores the benefits of progressively expanding the receptive field \citep{Ding_2022_CVPR} in iterative fusion for aggregating semantic information. Furthermore, guided by content density, our iterative fusion strategy tends to retain more informative speech frames \citep{ren-etal-2020-simulspeech}, resulting in the most significant performance improvement.

\subsection{Inference Efficiency}
\label{infer_efficiency}
\begin{wraptable}{r}{6.0cm}
\vspace{-0.1in} 
\caption{The inference efficiency on LibriTTS test subset of OpenASQA dataset, where ``Ours'' denotes FastLongSpeech.}
\small
\centering
\label{efficiency}
\begin{tabular}{c c c} \toprule[1.2pt]
\textbf{Method} & \textbf{Score} ($\uparrow$) & \textbf{TFLOPs} ($\downarrow$) \\ 

\cmidrule(lr){1-1} \cmidrule(lr){2-3}

Baseline & 3.73 & 9.79 \\
\cmidrule(lr){1-1} \cmidrule(lr){2-3}
Ours ($L$=400) & 3.80 & 8.54 \\
Ours ($L$=200) & 3.87 & 5.64 \\
Ours ($L$=100) & 3.71 & 4.17 \\

 \bottomrule[1pt]
\end{tabular}
\vspace{-0.2in} 
\end{wraptable}
After investigating the impact of various components in FastLongSpeech, we conduct analyses on the inference efficiency across different methods. To quantify this efficiency, we employ the TFLOPs metric, which measures the average number of floating-point operations (FLOPs) across the entire dataset and is calculated using calflops\footnote{\url{https://github.com/MrYxJ/calculate-flops.pytorch}} tool.  For long-speech scenarios, we incorporate average runtime as an additional efficiency indicator, which is measured in seconds. Table \ref{efficiency} and \ref{long_efficiency} present the results of inference efficiency experiments, which are obtained on NVIDIA L40.

In short-speech tasks, our approach demonstrates performance comparable to vanilla Qwen2-Audio while requiring only half the computational resources. 
When the allocated computational resources are increased, we can achieve better results. Notably, the computational costs decrease as the compression ratio increases. This not only demonstrates the better efficiency of our model but also highlights its ability to balance generation quality and inference efficiency.
\begin{wraptable}{r}{7.5cm}
\vspace{0.0in} 
\caption{The efficiency on the long-speech benchmark.}
\small
\centering
\label{long_efficiency}
\begin{tabular}{c c c c} \toprule[1.2pt]
\textbf{Method} & \textbf{Score} ($\uparrow$) & \textbf{TFLOPs} ($\downarrow$) & \textbf{Time} ($\downarrow$) \\ 

\cmidrule(lr){1-1} \cmidrule(lr){2-4}

NTK-RoPE & 3.44 & 61.21 & 4.80 \\

\cmidrule(lr){1-1} \cmidrule(lr){2-4}

Cascaded & \textbf{3.75} & n/a & 17.23+1.38 \\

\cmidrule(lr){1-1} \cmidrule(lr){2-4}
\textbf{Ours} & 3.55 & \textbf{26.44} & \textbf{1.47} \\

 \bottomrule[1pt]
\end{tabular}
\vspace{-0.1in} 
\end{wraptable}
The advantages of our method become even more pronounced in long-speech tasks, where our method achieves better generation quality than NTK-ROPE, with a 70\% reduction in runtime and a 60\% decrease in computational costs. Compared to the cascaded method, it even achieves a speedup of more than sevenfold, underscoring its substantial efficiency advantage for processing long-form speech. This further shows the effectiveness of our method in handling long-speech inputs. For spoken dialogue understanding, emotion recognition and speech information retrieval tasks, please refer to the Appendix \ref{vanilla_methods} and \ref{meld_section}.

\subsection{Content of Condensed Representations}
\begin{wraptable}{r}{5.8cm}
\vspace{-0.1in} 
\caption{The performance on the ASR task, where ``Ours'' denotes the FastLongSpeech. For the dataset, ``Clean'' and ``Other'' denote LibriSpeech test-clean and test-other sets. ``Giga'' denotes the test set of GigaSpeech. The results are evaluated with WER metric.}
\small
\centering
\label{asr}
\begin{tabular}{c c c c} \toprule[1.2pt]
\textbf{Method} & \textbf{Clean} & \textbf{Other} & \textbf{Giga} \\ 

\cmidrule(lr){1-1} \cmidrule(lr){2-4}

Baseline & 3.85 & 6.70 & 13.71 \\
\cmidrule(lr){1-1} \cmidrule(lr){2-4}
Ours ($L$=750) & 4.04 & 7.02 & 11.76 \\
Ours ($L$=400) & 4.08 & 7.17 & 11.77 \\
Ours ($L$=200) & 4.36 & 7.40 & 12.70 \\
Ours ($L$=100) & 27.12 & 24.61 & 23.69 \\

 \bottomrule[1pt]
\end{tabular}
\vspace{-0.1in} 
\end{wraptable}
Beyond the spoken QA, spoken dialogue understanding and emotion recognition tasks, we extend our evaluation to the ASR task, which requires precise transcription of the entire speech content \citep{10301513}. 
Through this task, we explore variations in condensed representations across different compression ratios. Table \ref{asr} demonstrates the ASR performance of Qwen2-Audio and our method. 
At low compression ratios ($L$=400), FastLongSpeech performs comparably to Qwen2-Audio, 
demonstrating the effectiveness of our dynamic compression training and iterative fusion strategy in preserving speech content. 
Unlike Qwen2-Audio, our method does not require substantial post-processing to extract the transcript, with strong instruction following abilities.
At higher compression ratios ($L$=100), FastLongSpeech slightly trails Qwen2-Audio in ASR but maintains comparable results in spoken QA, as illustrated in Figure \ref{main_res}. This indicates that while our approach demonstrates applicability across diverse tasks, the optimal compression ratio is inherently task-dependent. Therefore, achieving an effective balance between efficiency and effectiveness thus necessitates careful calibration and a thorough assessment of resource constraints.

\section{Related Work}
\paragraph{Large Speech-Language Models} With the advancements in Large Language Models (LLMs), recent research attempts to extend the understanding and reasoning capabilities of LLMs to speech inputs, becoming Large Speech-Language Models (LSLMs). Early studies \citep{shen2023hugginggptsolvingaitasks, huang2023audiogptunderstandinggeneratingspeech} employ a cascading paradigm, where speech is first transcribed into text before being processed by LLMs. More recently, some works \citep{chu2024qwen2audiotechnicalreport, fu2024vitaopensourceinteractiveomni, fang2024llamaomniseamlessspeechinteraction, microsoft2025phi4minitechnicalreportcompact} utilize the adaptors to align the output space of speech encoders with the input space of LLMs, achieving multi-task LSLMs. Other approaches \citep{rubenstein2023audiopalmlargelanguagemodel, zhang2023speechgptempoweringlargelanguage, défossez2024moshispeechtextfoundationmodel, zeng2024glm4voiceintelligenthumanlikeendtoend} utilize speech discretization techniques, converting waveforms into discrete units, enabling LSLMs to process speech in the same way they process text. These approaches allow LSLMs to handle both speech understanding and generation.

\paragraph{Long Sequence Modeling} 
Long sequence modeling presents challenges across diverse domains, including text, video, and speech. The approaches to long-context modeling vary depending on the type of the inputs. For extended text sequences, researchers explored methods such as position interpolation and extrapolation \citep{chen2023extendingcontextwindowlarge}, sliding window \citep{ratner-etal-2023-parallel}, continuous fine-tuning on long-text data \citep{rozière2024codellamaopenfoundation}, and native sparse attention \citep{yuan2025nativesparseattentionhardwarealigned}. 
To address long-video processing, recent works leverage frame selection or merging strategies \citep{song2024moviechatdensetokensparse}, as well as vision token merging techniques \citep{shang2024llavaprumergeadaptivetokenreduction}. In the realm of speech processing, early methods focus on enhancing the performance of ASR \citep{tsunoo2024decoderonlyarchitecturespeechrecognition} and speech translation \citep{gaido2021ctcbasedcompressiondirectspeech} through speech compression techniques. More recently, FastAdaSP \citep{lu2024fastadaspmultitaskadaptedefficientinference} mitigates inference overhead by performing token selection within LLMs. Concurrently, Speechprune \citep{lin2024speechprunecontextawaretokenpruning} employs a token selection strategy to extend the effective speech window of Qwen2-Audio to 90 seconds for Speech Information Retrieval task. StreamUni \citep{lin2025speechprunecontextawaretokenpruning} achieves real-time speech translation for long speech streams by integrating a segmentation strategy and a policy-decision module.

\section{Conclusion}
In this paper, we introduce FastLongSpeech, a novel approach that extends the capabilities of LSLMs to efficiently conduct long-speech processing. Experiments show that our method significantly reduces the computational costs and inference time in long-speech tasks, achieving better trade-offs between performance and efficiency.

\section*{Limitations}
Given the current scarcity of long-speech data, FastLongSpeech introduces an innovative dynamic compression training approach. This method leverages short-speech training data to extend the capabilities of LSLMs for long-speech processing. As long-speech training and evaluation data become more abundant in the future, FastLongSpeech will further enhance its ability to process longer speech inputs using the expanded datasets with lower costs.

\section*{Acknowledgement}
We gratefully acknowledge all the reviewers for their valuable comments and suggestions. This work was supported by the Natural Science Foundation of Beijing, China (Grant No. L257006).

\bibliographystyle{unsrtnat}
\bibliography{custom}

\begin{thebibliography}{58}
\providecommand{\natexlab}[1]{#1}
\providecommand{\url}[1]{\texttt{#1}}
\expandafter\ifx\csname urlstyle\endcsname\relax
  \providecommand{\doi}[1]{doi: #1}\else
  \providecommand{\doi}{doi: \begingroup \urlstyle{rm}\Url}\fi

\bibitem[OpenAI et~al.(2024)OpenAI, :, Hurst, Lerer, Goucher, Perelman, Ramesh, Clark, Ostrow, Welihinda, Hayes, Radford, Mądry, Baker-Whitcomb, Beutel, Borzunov, Carney, Chow, Kirillov, Nichol, Paino, Renzin, Passos, Kirillov, Christakis, Conneau, Kamali, Jabri, Moyer, Tam, Crookes, Tootoochian, Tootoonchian, Kumar, Vallone, Karpathy, Braunstein, Cann, Codispoti, Galu, Kondrich, Tulloch, Mishchenko, Baek, Jiang, Pelisse, Woodford, Gosalia, Dhar, Pantuliano, Nayak, Oliver, Zoph, Ghorbani, Leimberger, Rossen, Sokolowsky, Wang, Zweig, Hoover, Samic, McGrew, Spero, Giertler, Cheng, Lightcap, Walkin, Quinn, Guarraci, Hsu, Kellogg, Eastman, Lugaresi, Wainwright, Bassin, Hudson, Chu, Nelson, Li, Shern, Conger, Barette, Voss, Ding, Lu, Zhang, Beaumont, Hallacy, Koch, Gibson, Kim, Choi, McLeavey, Hesse, Fischer, Winter, Czarnecki, Jarvis, Wei, Koumouzelis, Sherburn, Kappler, Levin, Levy, Carr, Farhi, Mely, Robinson, Sasaki, Jin, Valladares, Tsipras, Li, Nguyen, Findlay, Oiwoh, Wong, Asdar, Proehl, Yang, Antonow,
  Kramer, Peterson, Sigler, Wallace, Brevdo, Mays, Khorasani, Such, Raso, Zhang, von Lohmann, Sulit, Goh, Oden, Salmon, Starace, Brockman, Salman, Bao, Hu, Wong, Wang, Schmidt, Whitney, Jun, Kirchner, de~Oliveira~Pinto, Ren, Chang, Chung, Kivlichan, O'Connell, O'Connell, Osband, Silber, Sohl, Okuyucu, Lan, Kostrikov, Sutskever, Kanitscheider, Gulrajani, Coxon, Menick, Pachocki, Aung, Betker, Crooks, Lennon, Kiros, Leike, Park, Kwon, Phang, Teplitz, Wei, Wolfe, Chen, Harris, Varavva, Lee, Shieh, Lin, Yu, Weng, Tang, Yu, Jang, Candela, Beutler, Landers, Parish, Heidecke, Schulman, Lachman, McKay, Uesato, Ward, Kim, Huizinga, Sitkin, Kraaijeveld, Gross, Kaplan, Snyder, Achiam, Jiao, Lee, Zhuang, Harriman, Fricke, Hayashi, Singhal, Shi, Karthik, Wood, Rimbach, Hsu, Nguyen, Gu-Lemberg, Button, Liu, Howe, Muthukumar, Luther, Ahmad, Kai, Itow, Workman, Pathak, Chen, Jing, Guy, Fedus, Zhou, Mamitsuka, Weng, McCallum, Held, Ouyang, Feuvrier, Zhang, Kondraciuk, Kaiser, Hewitt, Metz, Doshi, Aflak, Simens, Boyd,
  Thompson, Dukhan, Chen, Gray, Hudnall, Zhang, Aljubeh, Litwin, Zeng, Johnson, Shetty, Gupta, Shah, Yatbaz, Yang, Zhong, Glaese, Chen, Janner, Lampe, Petrov, Wu, Wang, Fradin, Pokrass, Castro, de~Castro, Pavlov, Brundage, Wang, Khan, Murati, Bavarian, Lin, Yesildal, Soto, Gimelshein, Cone, Staudacher, Summers, LaFontaine, Chowdhury, Ryder, Stathas, Turley, Tezak, Felix, Kudige, Keskar, Deutsch, Bundick, Puckett, Nachum, Okelola, Boiko, Murk, Jaffe, Watkins, Godement, Campbell-Moore, Chao, McMillan, Belov, Su, Bak, Bakkum, Deng, Dolan, Hoeschele, Welinder, Tillet, Pronin, Tillet, Dhariwal, Yuan, Dias, Lim, Arora, Troll, Lin, Lopes, Puri, Miyara, Leike, Gaubert, Zamani, Wang, Donnelly, Honsby, Smith, Sahai, Ramchandani, Huet, Carmichael, Zellers, Chen, Chen, Nigmatullin, Cheu, Jain, Altman, Schoenholz, Toizer, Miserendino, Agarwal, Culver, Ethersmith, Gray, Grove, Metzger, Hermani, Jain, Zhao, Wu, Jomoto, Wu, Shuaiqi, Xia, Phene, Papay, Narayanan, Coffey, Lee, Hall, Balaji, Broda, Stramer, Xu, Gogineni,
  Christianson, Sanders, Patwardhan, Cunninghman, Degry, Dimson, Raoux, Shadwell, Zheng, Underwood, Markov, Sherbakov, Rubin, Stasi, Kaftan, Heywood, Peterson, Walters, Eloundou, Qi, Moeller, Monaco, Kuo, Fomenko, Chang, Zheng, Zhou, Manassra, Sheu, Zaremba, Patil, Qian, Kim, Cheng, Zhang, He, Zhang, Jin, Dai, and Malkov]{openai2024gpt4ocard}
OpenAI, :, Aaron Hurst, Adam Lerer, Adam~P. Goucher, Adam Perelman, Aditya Ramesh, Aidan Clark, AJ~Ostrow, Akila Welihinda, Alan Hayes, Alec Radford, Aleksander Mądry, Alex Baker-Whitcomb, Alex Beutel, Alex Borzunov, Alex Carney, Alex Chow, Alex Kirillov, Alex Nichol, Alex Paino, Alex Renzin, Alex~Tachard Passos, Alexander Kirillov, Alexi Christakis, Alexis Conneau, Ali Kamali, Allan Jabri, Allison Moyer, Allison Tam, Amadou Crookes, Amin Tootoochian, Amin Tootoonchian, Ananya Kumar, Andrea Vallone, Andrej Karpathy, Andrew Braunstein, Andrew Cann, Andrew Codispoti, Andrew Galu, Andrew Kondrich, Andrew Tulloch, Andrey Mishchenko, Angela Baek, Angela Jiang, Antoine Pelisse, Antonia Woodford, Anuj Gosalia, Arka Dhar, Ashley Pantuliano, Avi Nayak, Avital Oliver, Barret Zoph, Behrooz Ghorbani, Ben Leimberger, Ben Rossen, Ben Sokolowsky, Ben Wang, Benjamin Zweig, Beth Hoover, Blake Samic, Bob McGrew, Bobby Spero, Bogo Giertler, Bowen Cheng, Brad Lightcap, Brandon Walkin, Brendan Quinn, Brian Guarraci, Brian Hsu,
  Bright Kellogg, Brydon Eastman, Camillo Lugaresi, Carroll Wainwright, Cary Bassin, Cary Hudson, Casey Chu, Chad Nelson, Chak Li, Chan~Jun Shern, Channing Conger, Charlotte Barette, Chelsea Voss, Chen Ding, Cheng Lu, Chong Zhang, Chris Beaumont, Chris Hallacy, Chris Koch, Christian Gibson, Christina Kim, Christine Choi, Christine McLeavey, Christopher Hesse, Claudia Fischer, Clemens Winter, Coley Czarnecki, Colin Jarvis, Colin Wei, Constantin Koumouzelis, Dane Sherburn, Daniel Kappler, Daniel Levin, Daniel Levy, David Carr, David Farhi, David Mely, David Robinson, David Sasaki, Denny Jin, Dev Valladares, Dimitris Tsipras, Doug Li, Duc~Phong Nguyen, Duncan Findlay, Edede Oiwoh, Edmund Wong, Ehsan Asdar, Elizabeth Proehl, Elizabeth Yang, Eric Antonow, Eric Kramer, Eric Peterson, Eric Sigler, Eric Wallace, Eugene Brevdo, Evan Mays, Farzad Khorasani, Felipe~Petroski Such, Filippo Raso, Francis Zhang, Fred von Lohmann, Freddie Sulit, Gabriel Goh, Gene Oden, Geoff Salmon, Giulio Starace, Greg Brockman, Hadi
  Salman, Haiming Bao, Haitang Hu, Hannah Wong, Haoyu Wang, Heather Schmidt, Heather Whitney, Heewoo Jun, Hendrik Kirchner, Henrique~Ponde de~Oliveira~Pinto, Hongyu Ren, Huiwen Chang, Hyung~Won Chung, Ian Kivlichan, Ian O'Connell, Ian O'Connell, Ian Osband, Ian Silber, Ian Sohl, Ibrahim Okuyucu, Ikai Lan, Ilya Kostrikov, Ilya Sutskever, Ingmar Kanitscheider, Ishaan Gulrajani, Jacob Coxon, Jacob Menick, Jakub Pachocki, James Aung, James Betker, James Crooks, James Lennon, Jamie Kiros, Jan Leike, Jane Park, Jason Kwon, Jason Phang, Jason Teplitz, Jason Wei, Jason Wolfe, Jay Chen, Jeff Harris, Jenia Varavva, Jessica~Gan Lee, Jessica Shieh, Ji~Lin, Jiahui Yu, Jiayi Weng, Jie Tang, Jieqi Yu, Joanne Jang, Joaquin~Quinonero Candela, Joe Beutler, Joe Landers, Joel Parish, Johannes Heidecke, John Schulman, Jonathan Lachman, Jonathan McKay, Jonathan Uesato, Jonathan Ward, Jong~Wook Kim, Joost Huizinga, Jordan Sitkin, Jos Kraaijeveld, Josh Gross, Josh Kaplan, Josh Snyder, Joshua Achiam, Joy Jiao, Joyce Lee, Juntang
  Zhuang, Justyn Harriman, Kai Fricke, Kai Hayashi, Karan Singhal, Katy Shi, Kavin Karthik, Kayla Wood, Kendra Rimbach, Kenny Hsu, Kenny Nguyen, Keren Gu-Lemberg, Kevin Button, Kevin Liu, Kiel Howe, Krithika Muthukumar, Kyle Luther, Lama Ahmad, Larry Kai, Lauren Itow, Lauren Workman, Leher Pathak, Leo Chen, Li~Jing, Lia Guy, Liam Fedus, Liang Zhou, Lien Mamitsuka, Lilian Weng, Lindsay McCallum, Lindsey Held, Long Ouyang, Louis Feuvrier, Lu~Zhang, Lukas Kondraciuk, Lukasz Kaiser, Luke Hewitt, Luke Metz, Lyric Doshi, Mada Aflak, Maddie Simens, Madelaine Boyd, Madeleine Thompson, Marat Dukhan, Mark Chen, Mark Gray, Mark Hudnall, Marvin Zhang, Marwan Aljubeh, Mateusz Litwin, Matthew Zeng, Max Johnson, Maya Shetty, Mayank Gupta, Meghan Shah, Mehmet Yatbaz, Meng~Jia Yang, Mengchao Zhong, Mia Glaese, Mianna Chen, Michael Janner, Michael Lampe, Michael Petrov, Michael Wu, Michele Wang, Michelle Fradin, Michelle Pokrass, Miguel Castro, Miguel Oom~Temudo de~Castro, Mikhail Pavlov, Miles Brundage, Miles Wang, Minal
  Khan, Mira Murati, Mo~Bavarian, Molly Lin, Murat Yesildal, Nacho Soto, Natalia Gimelshein, Natalie Cone, Natalie Staudacher, Natalie Summers, Natan LaFontaine, Neil Chowdhury, Nick Ryder, Nick Stathas, Nick Turley, Nik Tezak, Niko Felix, Nithanth Kudige, Nitish Keskar, Noah Deutsch, Noel Bundick, Nora Puckett, Ofir Nachum, Ola Okelola, Oleg Boiko, Oleg Murk, Oliver Jaffe, Olivia Watkins, Olivier Godement, Owen Campbell-Moore, Patrick Chao, Paul McMillan, Pavel Belov, Peng Su, Peter Bak, Peter Bakkum, Peter Deng, Peter Dolan, Peter Hoeschele, Peter Welinder, Phil Tillet, Philip Pronin, Philippe Tillet, Prafulla Dhariwal, Qiming Yuan, Rachel Dias, Rachel Lim, Rahul Arora, Rajan Troll, Randall Lin, Rapha~Gontijo Lopes, Raul Puri, Reah Miyara, Reimar Leike, Renaud Gaubert, Reza Zamani, Ricky Wang, Rob Donnelly, Rob Honsby, Rocky Smith, Rohan Sahai, Rohit Ramchandani, Romain Huet, Rory Carmichael, Rowan Zellers, Roy Chen, Ruby Chen, Ruslan Nigmatullin, Ryan Cheu, Saachi Jain, Sam Altman, Sam Schoenholz, Sam
  Toizer, Samuel Miserendino, Sandhini Agarwal, Sara Culver, Scott Ethersmith, Scott Gray, Sean Grove, Sean Metzger, Shamez Hermani, Shantanu Jain, Shengjia Zhao, Sherwin Wu, Shino Jomoto, Shirong Wu, Shuaiqi, Xia, Sonia Phene, Spencer Papay, Srinivas Narayanan, Steve Coffey, Steve Lee, Stewart Hall, Suchir Balaji, Tal Broda, Tal Stramer, Tao Xu, Tarun Gogineni, Taya Christianson, Ted Sanders, Tejal Patwardhan, Thomas Cunninghman, Thomas Degry, Thomas Dimson, Thomas Raoux, Thomas Shadwell, Tianhao Zheng, Todd Underwood, Todor Markov, Toki Sherbakov, Tom Rubin, Tom Stasi, Tomer Kaftan, Tristan Heywood, Troy Peterson, Tyce Walters, Tyna Eloundou, Valerie Qi, Veit Moeller, Vinnie Monaco, Vishal Kuo, Vlad Fomenko, Wayne Chang, Weiyi Zheng, Wenda Zhou, Wesam Manassra, Will Sheu, Wojciech Zaremba, Yash Patil, Yilei Qian, Yongjik Kim, Youlong Cheng, Yu~Zhang, Yuchen He, Yuchen Zhang, Yujia Jin, Yunxing Dai, and Yury Malkov.
\newblock Gpt-4o system card, 2024.
\newblock URL \url{https://arxiv.org/abs/2410.21276}.

\bibitem[Grattafiori et~al.(2024)Grattafiori, Dubey, Jauhri, Pandey, Kadian, Al-Dahle, Letman, Mathur, Schelten, Vaughan, Yang, Fan, Goyal, Hartshorn, Yang, Mitra, Sravankumar, Korenev, Hinsvark, Rao, Zhang, Rodriguez, Gregerson, Spataru, Roziere, Biron, Tang, Chern, Caucheteux, Nayak, Bi, Marra, McConnell, Keller, Touret, Wu, Wong, Ferrer, Nikolaidis, Allonsius, Song, Pintz, Livshits, Wyatt, Esiobu, Choudhary, Mahajan, Garcia-Olano, Perino, Hupkes, Lakomkin, AlBadawy, Lobanova, Dinan, Smith, Radenovic, Guzmán, Zhang, Synnaeve, Lee, Anderson, Thattai, Nail, Mialon, Pang, Cucurell, Nguyen, Korevaar, Xu, Touvron, Zarov, Ibarra, Kloumann, Misra, Evtimov, Zhang, Copet, Lee, Geffert, Vranes, Park, Mahadeokar, Shah, van~der Linde, Billock, Hong, Lee, Fu, Chi, Huang, Liu, Wang, Yu, Bitton, Spisak, Park, Rocca, Johnstun, Saxe, Jia, Alwala, Prasad, Upasani, Plawiak, Li, Heafield, Stone, El-Arini, Iyer, Malik, Chiu, Bhalla, Lakhotia, Rantala-Yeary, van~der Maaten, Chen, Tan, Jenkins, Martin, Madaan, Malo, Blecher,
  Landzaat, de~Oliveira, Muzzi, Pasupuleti, Singh, Paluri, Kardas, Tsimpoukelli, Oldham, Rita, Pavlova, Kambadur, Lewis, Si, Singh, Hassan, Goyal, Torabi, Bashlykov, Bogoychev, Chatterji, Zhang, Duchenne, Çelebi, Alrassy, Zhang, Li, Vasic, Weng, Bhargava, Dubal, Krishnan, Koura, Xu, He, Dong, Srinivasan, Ganapathy, Calderer, Cabral, Stojnic, Raileanu, Maheswari, Girdhar, Patel, Sauvestre, Polidoro, Sumbaly, Taylor, Silva, Hou, Wang, Hosseini, Chennabasappa, Singh, Bell, Kim, Edunov, Nie, Narang, Raparthy, Shen, Wan, Bhosale, Zhang, Vandenhende, Batra, Whitman, Sootla, Collot, Gururangan, Borodinsky, Herman, Fowler, Sheasha, Georgiou, Scialom, Speckbacher, Mihaylov, Xiao, Karn, Goswami, Gupta, Ramanathan, Kerkez, Gonguet, Do, Vogeti, Albiero, Petrovic, Chu, Xiong, Fu, Meers, Martinet, Wang, Wang, Tan, Xia, Xie, Jia, Wang, Goldschlag, Gaur, Babaei, Wen, Song, Zhang, Li, Mao, Coudert, Yan, Chen, Papakipos, Singh, Srivastava, Jain, Kelsey, Shajnfeld, Gangidi, Victoria, Goldstand, Menon, Sharma, Boesenberg,
  Baevski, Feinstein, Kallet, Sangani, Teo, Yunus, Lupu, Alvarado, Caples, Gu, Ho, Poulton, Ryan, Ramchandani, Dong, Franco, Goyal, Saraf, Chowdhury, Gabriel, Bharambe, Eisenman, Yazdan, James, Maurer, Leonhardi, Huang, Loyd, Paola, Paranjape, Liu, Wu, Ni, Hancock, Wasti, Spence, Stojkovic, Gamido, Montalvo, Parker, Burton, Mejia, Liu, Wang, Kim, Zhou, Hu, Chu, Cai, Tindal, Feichtenhofer, Gao, Civin, Beaty, Kreymer, Li, Adkins, Xu, Testuggine, David, Parikh, Liskovich, Foss, Wang, Le, Holland, Dowling, Jamil, Montgomery, Presani, Hahn, Wood, Le, Brinkman, Arcaute, Dunbar, Smothers, Sun, Kreuk, Tian, Kokkinos, Ozgenel, Caggioni, Kanayet, Seide, Florez, Schwarz, Badeer, Swee, Halpern, Herman, Sizov, Guangyi, Zhang, Lakshminarayanan, Inan, Shojanazeri, Zou, Wang, Zha, Habeeb, Rudolph, Suk, Aspegren, Goldman, Zhan, Damlaj, Molybog, Tufanov, Leontiadis, Veliche, Gat, Weissman, Geboski, Kohli, Lam, Asher, Gaya, Marcus, Tang, Chan, Zhen, Reizenstein, Teboul, Zhong, Jin, Yang, Cummings, Carvill, Shepard, McPhie,
  Torres, Ginsburg, Wang, Wu, U, Saxena, Khandelwal, Zand, Matosich, Veeraraghavan, Michelena, Li, Jagadeesh, Huang, Chawla, Huang, Chen, Garg, A, Silva, Bell, Zhang, Guo, Yu, Moshkovich, Wehrstedt, Khabsa, Avalani, Bhatt, Mankus, Hasson, Lennie, Reso, Groshev, Naumov, Lathi, Keneally, Liu, Seltzer, Valko, Restrepo, Patel, Vyatskov, Samvelyan, Clark, Macey, Wang, Hermoso, Metanat, Rastegari, Bansal, Santhanam, Parks, White, Bawa, Singhal, Egebo, Usunier, Mehta, Laptev, Dong, Cheng, Chernoguz, Hart, Salpekar, Kalinli, Kent, Parekh, Saab, Balaji, Rittner, Bontrager, Roux, Dollar, Zvyagina, Ratanchandani, Yuvraj, Liang, Alao, Rodriguez, Ayub, Murthy, Nayani, Mitra, Parthasarathy, Li, Hogan, Battey, Wang, Howes, Rinott, Mehta, Siby, Bondu, Datta, Chugh, Hunt, Dhillon, Sidorov, Pan, Mahajan, Verma, Yamamoto, Ramaswamy, Lindsay, Lindsay, Feng, Lin, Zha, Patil, Shankar, Zhang, Zhang, Wang, Agarwal, Sajuyigbe, Chintala, Max, Chen, Kehoe, Satterfield, Govindaprasad, Gupta, Deng, Cho, Virk, Subramanian, Choudhury,
  Goldman, Remez, Glaser, Best, Koehler, Robinson, Li, Zhang, Matthews, Chou, Shaked, Vontimitta, Ajayi, Montanez, Mohan, Kumar, Mangla, Ionescu, Poenaru, Mihailescu, Ivanov, Li, Wang, Jiang, Bouaziz, Constable, Tang, Wu, Wang, Wu, Gao, Kleinman, Chen, Hu, Jia, Qi, Li, Zhang, Zhang, Adi, Nam, Yu, Wang, Zhao, Hao, Qian, Li, He, Rait, DeVito, Rosnbrick, Wen, Yang, Zhao, and Ma]{grattafiori2024llama3herdmodels}
Aaron Grattafiori, Abhimanyu Dubey, Abhinav Jauhri, Abhinav Pandey, Abhishek Kadian, Ahmad Al-Dahle, Aiesha Letman, Akhil Mathur, Alan Schelten, Alex Vaughan, Amy Yang, Angela Fan, Anirudh Goyal, Anthony Hartshorn, Aobo Yang, Archi Mitra, Archie Sravankumar, Artem Korenev, Arthur Hinsvark, Arun Rao, Aston Zhang, Aurelien Rodriguez, Austen Gregerson, Ava Spataru, Baptiste Roziere, Bethany Biron, Binh Tang, Bobbie Chern, Charlotte Caucheteux, Chaya Nayak, Chloe Bi, Chris Marra, Chris McConnell, Christian Keller, Christophe Touret, Chunyang Wu, Corinne Wong, Cristian~Canton Ferrer, Cyrus Nikolaidis, Damien Allonsius, Daniel Song, Danielle Pintz, Danny Livshits, Danny Wyatt, David Esiobu, Dhruv Choudhary, Dhruv Mahajan, Diego Garcia-Olano, Diego Perino, Dieuwke Hupkes, Egor Lakomkin, Ehab AlBadawy, Elina Lobanova, Emily Dinan, Eric~Michael Smith, Filip Radenovic, Francisco Guzmán, Frank Zhang, Gabriel Synnaeve, Gabrielle Lee, Georgia~Lewis Anderson, Govind Thattai, Graeme Nail, Gregoire Mialon, Guan Pang,
  Guillem Cucurell, Hailey Nguyen, Hannah Korevaar, Hu~Xu, Hugo Touvron, Iliyan Zarov, Imanol~Arrieta Ibarra, Isabel Kloumann, Ishan Misra, Ivan Evtimov, Jack Zhang, Jade Copet, Jaewon Lee, Jan Geffert, Jana Vranes, Jason Park, Jay Mahadeokar, Jeet Shah, Jelmer van~der Linde, Jennifer Billock, Jenny Hong, Jenya Lee, Jeremy Fu, Jianfeng Chi, Jianyu Huang, Jiawen Liu, Jie Wang, Jiecao Yu, Joanna Bitton, Joe Spisak, Jongsoo Park, Joseph Rocca, Joshua Johnstun, Joshua Saxe, Junteng Jia, Kalyan~Vasuden Alwala, Karthik Prasad, Kartikeya Upasani, Kate Plawiak, Ke~Li, Kenneth Heafield, Kevin Stone, Khalid El-Arini, Krithika Iyer, Kshitiz Malik, Kuenley Chiu, Kunal Bhalla, Kushal Lakhotia, Lauren Rantala-Yeary, Laurens van~der Maaten, Lawrence Chen, Liang Tan, Liz Jenkins, Louis Martin, Lovish Madaan, Lubo Malo, Lukas Blecher, Lukas Landzaat, Luke de~Oliveira, Madeline Muzzi, Mahesh Pasupuleti, Mannat Singh, Manohar Paluri, Marcin Kardas, Maria Tsimpoukelli, Mathew Oldham, Mathieu Rita, Maya Pavlova, Melanie Kambadur,
  Mike Lewis, Min Si, Mitesh~Kumar Singh, Mona Hassan, Naman Goyal, Narjes Torabi, Nikolay Bashlykov, Nikolay Bogoychev, Niladri Chatterji, Ning Zhang, Olivier Duchenne, Onur Çelebi, Patrick Alrassy, Pengchuan Zhang, Pengwei Li, Petar Vasic, Peter Weng, Prajjwal Bhargava, Pratik Dubal, Praveen Krishnan, Punit~Singh Koura, Puxin Xu, Qing He, Qingxiao Dong, Ragavan Srinivasan, Raj Ganapathy, Ramon Calderer, Ricardo~Silveira Cabral, Robert Stojnic, Roberta Raileanu, Rohan Maheswari, Rohit Girdhar, Rohit Patel, Romain Sauvestre, Ronnie Polidoro, Roshan Sumbaly, Ross Taylor, Ruan Silva, Rui Hou, Rui Wang, Saghar Hosseini, Sahana Chennabasappa, Sanjay Singh, Sean Bell, Seohyun~Sonia Kim, Sergey Edunov, Shaoliang Nie, Sharan Narang, Sharath Raparthy, Sheng Shen, Shengye Wan, Shruti Bhosale, Shun Zhang, Simon Vandenhende, Soumya Batra, Spencer Whitman, Sten Sootla, Stephane Collot, Suchin Gururangan, Sydney Borodinsky, Tamar Herman, Tara Fowler, Tarek Sheasha, Thomas Georgiou, Thomas Scialom, Tobias Speckbacher,
  Todor Mihaylov, Tong Xiao, Ujjwal Karn, Vedanuj Goswami, Vibhor Gupta, Vignesh Ramanathan, Viktor Kerkez, Vincent Gonguet, Virginie Do, Vish Vogeti, Vítor Albiero, Vladan Petrovic, Weiwei Chu, Wenhan Xiong, Wenyin Fu, Whitney Meers, Xavier Martinet, Xiaodong Wang, Xiaofang Wang, Xiaoqing~Ellen Tan, Xide Xia, Xinfeng Xie, Xuchao Jia, Xuewei Wang, Yaelle Goldschlag, Yashesh Gaur, Yasmine Babaei, Yi~Wen, Yiwen Song, Yuchen Zhang, Yue Li, Yuning Mao, Zacharie~Delpierre Coudert, Zheng Yan, Zhengxing Chen, Zoe Papakipos, Aaditya Singh, Aayushi Srivastava, Abha Jain, Adam Kelsey, Adam Shajnfeld, Adithya Gangidi, Adolfo Victoria, Ahuva Goldstand, Ajay Menon, Ajay Sharma, Alex Boesenberg, Alexei Baevski, Allie Feinstein, Amanda Kallet, Amit Sangani, Amos Teo, Anam Yunus, Andrei Lupu, Andres Alvarado, Andrew Caples, Andrew Gu, Andrew Ho, Andrew Poulton, Andrew Ryan, Ankit Ramchandani, Annie Dong, Annie Franco, Anuj Goyal, Aparajita Saraf, Arkabandhu Chowdhury, Ashley Gabriel, Ashwin Bharambe, Assaf Eisenman, Azadeh
  Yazdan, Beau James, Ben Maurer, Benjamin Leonhardi, Bernie Huang, Beth Loyd, Beto~De Paola, Bhargavi Paranjape, Bing Liu, Bo~Wu, Boyu Ni, Braden Hancock, Bram Wasti, Brandon Spence, Brani Stojkovic, Brian Gamido, Britt Montalvo, Carl Parker, Carly Burton, Catalina Mejia, Ce~Liu, Changhan Wang, Changkyu Kim, Chao Zhou, Chester Hu, Ching-Hsiang Chu, Chris Cai, Chris Tindal, Christoph Feichtenhofer, Cynthia Gao, Damon Civin, Dana Beaty, Daniel Kreymer, Daniel Li, David Adkins, David Xu, Davide Testuggine, Delia David, Devi Parikh, Diana Liskovich, Didem Foss, Dingkang Wang, Duc Le, Dustin Holland, Edward Dowling, Eissa Jamil, Elaine Montgomery, Eleonora Presani, Emily Hahn, Emily Wood, Eric-Tuan Le, Erik Brinkman, Esteban Arcaute, Evan Dunbar, Evan Smothers, Fei Sun, Felix Kreuk, Feng Tian, Filippos Kokkinos, Firat Ozgenel, Francesco Caggioni, Frank Kanayet, Frank Seide, Gabriela~Medina Florez, Gabriella Schwarz, Gada Badeer, Georgia Swee, Gil Halpern, Grant Herman, Grigory Sizov, Guangyi, Zhang, Guna
  Lakshminarayanan, Hakan Inan, Hamid Shojanazeri, Han Zou, Hannah Wang, Hanwen Zha, Haroun Habeeb, Harrison Rudolph, Helen Suk, Henry Aspegren, Hunter Goldman, Hongyuan Zhan, Ibrahim Damlaj, Igor Molybog, Igor Tufanov, Ilias Leontiadis, Irina-Elena Veliche, Itai Gat, Jake Weissman, James Geboski, James Kohli, Janice Lam, Japhet Asher, Jean-Baptiste Gaya, Jeff Marcus, Jeff Tang, Jennifer Chan, Jenny Zhen, Jeremy Reizenstein, Jeremy Teboul, Jessica Zhong, Jian Jin, Jingyi Yang, Joe Cummings, Jon Carvill, Jon Shepard, Jonathan McPhie, Jonathan Torres, Josh Ginsburg, Junjie Wang, Kai Wu, Kam~Hou U, Karan Saxena, Kartikay Khandelwal, Katayoun Zand, Kathy Matosich, Kaushik Veeraraghavan, Kelly Michelena, Keqian Li, Kiran Jagadeesh, Kun Huang, Kunal Chawla, Kyle Huang, Lailin Chen, Lakshya Garg, Lavender A, Leandro Silva, Lee Bell, Lei Zhang, Liangpeng Guo, Licheng Yu, Liron Moshkovich, Luca Wehrstedt, Madian Khabsa, Manav Avalani, Manish Bhatt, Martynas Mankus, Matan Hasson, Matthew Lennie, Matthias Reso, Maxim
  Groshev, Maxim Naumov, Maya Lathi, Meghan Keneally, Miao Liu, Michael~L. Seltzer, Michal Valko, Michelle Restrepo, Mihir Patel, Mik Vyatskov, Mikayel Samvelyan, Mike Clark, Mike Macey, Mike Wang, Miquel~Jubert Hermoso, Mo~Metanat, Mohammad Rastegari, Munish Bansal, Nandhini Santhanam, Natascha Parks, Natasha White, Navyata Bawa, Nayan Singhal, Nick Egebo, Nicolas Usunier, Nikhil Mehta, Nikolay~Pavlovich Laptev, Ning Dong, Norman Cheng, Oleg Chernoguz, Olivia Hart, Omkar Salpekar, Ozlem Kalinli, Parkin Kent, Parth Parekh, Paul Saab, Pavan Balaji, Pedro Rittner, Philip Bontrager, Pierre Roux, Piotr Dollar, Polina Zvyagina, Prashant Ratanchandani, Pritish Yuvraj, Qian Liang, Rachad Alao, Rachel Rodriguez, Rafi Ayub, Raghotham Murthy, Raghu Nayani, Rahul Mitra, Rangaprabhu Parthasarathy, Raymond Li, Rebekkah Hogan, Robin Battey, Rocky Wang, Russ Howes, Ruty Rinott, Sachin Mehta, Sachin Siby, Sai~Jayesh Bondu, Samyak Datta, Sara Chugh, Sara Hunt, Sargun Dhillon, Sasha Sidorov, Satadru Pan, Saurabh Mahajan,
  Saurabh Verma, Seiji Yamamoto, Sharadh Ramaswamy, Shaun Lindsay, Shaun Lindsay, Sheng Feng, Shenghao Lin, Shengxin~Cindy Zha, Shishir Patil, Shiva Shankar, Shuqiang Zhang, Shuqiang Zhang, Sinong Wang, Sneha Agarwal, Soji Sajuyigbe, Soumith Chintala, Stephanie Max, Stephen Chen, Steve Kehoe, Steve Satterfield, Sudarshan Govindaprasad, Sumit Gupta, Summer Deng, Sungmin Cho, Sunny Virk, Suraj Subramanian, Sy~Choudhury, Sydney Goldman, Tal Remez, Tamar Glaser, Tamara Best, Thilo Koehler, Thomas Robinson, Tianhe Li, Tianjun Zhang, Tim Matthews, Timothy Chou, Tzook Shaked, Varun Vontimitta, Victoria Ajayi, Victoria Montanez, Vijai Mohan, Vinay~Satish Kumar, Vishal Mangla, Vlad Ionescu, Vlad Poenaru, Vlad~Tiberiu Mihailescu, Vladimir Ivanov, Wei Li, Wenchen Wang, Wenwen Jiang, Wes Bouaziz, Will Constable, Xiaocheng Tang, Xiaojian Wu, Xiaolan Wang, Xilun Wu, Xinbo Gao, Yaniv Kleinman, Yanjun Chen, Ye~Hu, Ye~Jia, Ye~Qi, Yenda Li, Yilin Zhang, Ying Zhang, Yossi Adi, Youngjin Nam, Yu, Wang, Yu~Zhao, Yuchen Hao, Yundi
  Qian, Yunlu Li, Yuzi He, Zach Rait, Zachary DeVito, Zef Rosnbrick, Zhaoduo Wen, Zhenyu Yang, Zhiwei Zhao, and Zhiyu Ma.
\newblock The llama 3 herd of models, 2024.
\newblock URL \url{https://arxiv.org/abs/2407.21783}.

\bibitem[DeepSeek-AI et~al.(2024)DeepSeek-AI, Liu, Feng, Xue, Wang, Wu, Lu, Zhao, Deng, Zhang, Ruan, Dai, Guo, Yang, Chen, Ji, Li, Lin, Dai, Luo, Hao, Chen, Li, Zhang, Bao, Xu, Wang, Zhang, Ding, Xin, Gao, Li, Qu, Cai, Liang, Guo, Ni, Li, Wang, Chen, Chen, Yuan, Qiu, Li, Song, Dong, Hu, Gao, Guan, Huang, Yu, Wang, Zhang, Xu, Xia, Zhao, Wang, Zhang, Li, Wang, Zhang, Zhang, Tang, Li, Tian, Huang, Wang, Zhang, Wang, Zhu, Chen, Du, Chen, Jin, Ge, Zhang, Pan, Wang, Xu, Zhang, Chen, Li, Lu, Zhou, Chen, Wu, Ye, Ye, Ma, Wang, Zhou, Yu, Zhou, Pan, Wang, Yun, Pei, Sun, Xiao, Zeng, Zhao, An, Liu, Liang, Gao, Yu, Zhang, Li, Jin, Wang, Bi, Liu, Wang, Shen, Chen, Zhang, Chen, Nie, Sun, Wang, Cheng, Liu, Xie, Liu, Yu, Song, Shan, Zhou, Yang, Li, Su, Lin, Li, Wang, Wei, Zhu, Zhang, Xu, Xu, Huang, Li, Zhao, Sun, Li, Wang, Yu, Zheng, Zhang, Shi, Xiong, He, Tang, Piao, Wang, Tan, Ma, Liu, Guo, Wu, Ou, Zhu, Wang, Gong, Zou, He, Zha, Xiong, Ma, Yan, Luo, You, Liu, Zhou, Wu, Ren, Ren, Sha, Fu, Xu, Huang, Zhang, Xie, Zhang, Hao,
  Gou, Ma, Yan, Shao, Xu, Wu, Zhang, Li, Gu, Zhu, Liu, Li, Xie, Song, Gao, and Pan]{deepseekai2024deepseekv3technicalreport}
DeepSeek-AI, Aixin Liu, Bei Feng, Bing Xue, Bingxuan Wang, Bochao Wu, Chengda Lu, Chenggang Zhao, Chengqi Deng, Chenyu Zhang, Chong Ruan, Damai Dai, Daya Guo, Dejian Yang, Deli Chen, Dongjie Ji, Erhang Li, Fangyun Lin, Fucong Dai, Fuli Luo, Guangbo Hao, Guanting Chen, Guowei Li, H.~Zhang, Han Bao, Hanwei Xu, Haocheng Wang, Haowei Zhang, Honghui Ding, Huajian Xin, Huazuo Gao, Hui Li, Hui Qu, J.~L. Cai, Jian Liang, Jianzhong Guo, Jiaqi Ni, Jiashi Li, Jiawei Wang, Jin Chen, Jingchang Chen, Jingyang Yuan, Junjie Qiu, Junlong Li, Junxiao Song, Kai Dong, Kai Hu, Kaige Gao, Kang Guan, Kexin Huang, Kuai Yu, Lean Wang, Lecong Zhang, Lei Xu, Leyi Xia, Liang Zhao, Litong Wang, Liyue Zhang, Meng Li, Miaojun Wang, Mingchuan Zhang, Minghua Zhang, Minghui Tang, Mingming Li, Ning Tian, Panpan Huang, Peiyi Wang, Peng Zhang, Qiancheng Wang, Qihao Zhu, Qinyu Chen, Qiushi Du, R.~J. Chen, R.~L. Jin, Ruiqi Ge, Ruisong Zhang, Ruizhe Pan, Runji Wang, Runxin Xu, Ruoyu Zhang, Ruyi Chen, S.~S. Li, Shanghao Lu, Shangyan Zhou, Shanhuang
  Chen, Shaoqing Wu, Shengfeng Ye, Shengfeng Ye, Shirong Ma, Shiyu Wang, Shuang Zhou, Shuiping Yu, Shunfeng Zhou, Shuting Pan, T.~Wang, Tao Yun, Tian Pei, Tianyu Sun, W.~L. Xiao, Wangding Zeng, Wanjia Zhao, Wei An, Wen Liu, Wenfeng Liang, Wenjun Gao, Wenqin Yu, Wentao Zhang, X.~Q. Li, Xiangyue Jin, Xianzu Wang, Xiao Bi, Xiaodong Liu, Xiaohan Wang, Xiaojin Shen, Xiaokang Chen, Xiaokang Zhang, Xiaosha Chen, Xiaotao Nie, Xiaowen Sun, Xiaoxiang Wang, Xin Cheng, Xin Liu, Xin Xie, Xingchao Liu, Xingkai Yu, Xinnan Song, Xinxia Shan, Xinyi Zhou, Xinyu Yang, Xinyuan Li, Xuecheng Su, Xuheng Lin, Y.~K. Li, Y.~Q. Wang, Y.~X. Wei, Y.~X. Zhu, Yang Zhang, Yanhong Xu, Yanhong Xu, Yanping Huang, Yao Li, Yao Zhao, Yaofeng Sun, Yaohui Li, Yaohui Wang, Yi~Yu, Yi~Zheng, Yichao Zhang, Yifan Shi, Yiliang Xiong, Ying He, Ying Tang, Yishi Piao, Yisong Wang, Yixuan Tan, Yiyang Ma, Yiyuan Liu, Yongqiang Guo, Yu~Wu, Yuan Ou, Yuchen Zhu, Yuduan Wang, Yue Gong, Yuheng Zou, Yujia He, Yukun Zha, Yunfan Xiong, Yunxian Ma, Yuting Yan, Yuxiang
  Luo, Yuxiang You, Yuxuan Liu, Yuyang Zhou, Z.~F. Wu, Z.~Z. Ren, Zehui Ren, Zhangli Sha, Zhe Fu, Zhean Xu, Zhen Huang, Zhen Zhang, Zhenda Xie, Zhengyan Zhang, Zhewen Hao, Zhibin Gou, Zhicheng Ma, Zhigang Yan, Zhihong Shao, Zhipeng Xu, Zhiyu Wu, Zhongyu Zhang, Zhuoshu Li, Zihui Gu, Zijia Zhu, Zijun Liu, Zilin Li, Ziwei Xie, Ziyang Song, Ziyi Gao, and Zizheng Pan.
\newblock Deepseek-v3 technical report, 2024.
\newblock URL \url{https://arxiv.org/abs/2412.19437}.

\bibitem[Tang et~al.(2024)Tang, Yu, Sun, Chen, Tan, Li, Lu, MA, and Zhang]{tang2024salmonn}
Changli Tang, Wenyi Yu, Guangzhi Sun, Xianzhao Chen, Tian Tan, Wei Li, Lu~Lu, Zejun MA, and Chao Zhang.
\newblock {SALMONN}: Towards generic hearing abilities for large language models.
\newblock In \emph{The Twelfth International Conference on Learning Representations}, 2024.
\newblock URL \url{https://openreview.net/forum?id=14rn7HpKVk}.

\bibitem[Chu et~al.(2024)Chu, Xu, Yang, Wei, Wei, Guo, Leng, Lv, He, Lin, Zhou, and Zhou]{chu2024qwen2audiotechnicalreport}
Yunfei Chu, Jin Xu, Qian Yang, Haojie Wei, Xipin Wei, Zhifang Guo, Yichong Leng, Yuanjun Lv, Jinzheng He, Junyang Lin, Chang Zhou, and Jingren Zhou.
\newblock Qwen2-audio technical report, 2024.
\newblock URL \url{https://arxiv.org/abs/2407.10759}.

\bibitem[Chen et~al.(2025)Chen, Chen, Chen, Chen, Chen, Deng, Du, Gao, Gao, Gao, Li, Lv, Liu, Luo, Ma, Ni, Shi, Tang, Wang, Wang, Wang, Wang, Xu, Yu, Yan, Yang, Yang, Yang, Yang, Zhao, Zhang, Zhang, Zhao, Zhang, Zhang, and Zhou]{chen2025minmomultimodallargelanguage}
Qian Chen, Yafeng Chen, Yanni Chen, Mengzhe Chen, Yingda Chen, Chong Deng, Zhihao Du, Ruize Gao, Changfeng Gao, Zhifu Gao, Yabin Li, Xiang Lv, Jiaqing Liu, Haoneng Luo, Bin Ma, Chongjia Ni, Xian Shi, Jialong Tang, Hui Wang, Hao Wang, Wen Wang, Yuxuan Wang, Yunlan Xu, Fan Yu, Zhijie Yan, Yexin Yang, Baosong Yang, Xian Yang, Guanrou Yang, Tianyu Zhao, Qinglin Zhang, Shiliang Zhang, Nan Zhao, Pei Zhang, Chong Zhang, and Jinren Zhou.
\newblock Minmo: A multimodal large language model for seamless voice interaction, 2025.
\newblock URL \url{https://arxiv.org/abs/2501.06282}.

\bibitem[Chu et~al.(2023)Chu, Xu, Zhou, Yang, Zhang, Yan, Zhou, and Zhou]{chu2023qwenaudioadvancinguniversalaudio}
Yunfei Chu, Jin Xu, Xiaohuan Zhou, Qian Yang, Shiliang Zhang, Zhijie Yan, Chang Zhou, and Jingren Zhou.
\newblock Qwen-audio: Advancing universal audio understanding via unified large-scale audio-language models, 2023.
\newblock URL \url{https://arxiv.org/abs/2311.07919}.

\bibitem[Shen et~al.(2023)Shen, Song, Tan, Li, Lu, and Zhuang]{shen2023hugginggptsolvingaitasks}
Yongliang Shen, Kaitao Song, Xu~Tan, Dongsheng Li, Weiming Lu, and Yueting Zhuang.
\newblock Hugginggpt: Solving ai tasks with chatgpt and its friends in hugging face, 2023.
\newblock URL \url{https://arxiv.org/abs/2303.17580}.

\bibitem[Huang et~al.(2023)Huang, Li, Yang, Shi, Chang, Ye, Wu, Hong, Huang, Liu, Ren, Zhao, and Watanabe]{huang2023audiogptunderstandinggeneratingspeech}
Rongjie Huang, Mingze Li, Dongchao Yang, Jiatong Shi, Xuankai Chang, Zhenhui Ye, Yuning Wu, Zhiqing Hong, Jiawei Huang, Jinglin Liu, Yi~Ren, Zhou Zhao, and Shinji Watanabe.
\newblock Audiogpt: Understanding and generating speech, music, sound, and talking head, 2023.
\newblock URL \url{https://arxiv.org/abs/2304.12995}.

\bibitem[Wang et~al.(2024)Wang, Liao, Huang, Lu, Wu, Liu, Zong, and Zhang]{wang2024blspbootstrappinglanguagespeechpretraining}
Chen Wang, Minpeng Liao, Zhongqiang Huang, Jinliang Lu, Junhong Wu, Yuchen Liu, Chengqing Zong, and Jiajun Zhang.
\newblock Blsp: Bootstrapping language-speech pre-training via behavior alignment of continuation writing, 2024.
\newblock URL \url{https://arxiv.org/abs/2309.00916}.

\bibitem[Zhang et~al.(2023)Zhang, Li, Zhang, Zhan, Wang, Zhou, and Qiu]{zhang2023speechgptempoweringlargelanguage}
Dong Zhang, Shimin Li, Xin Zhang, Jun Zhan, Pengyu Wang, Yaqian Zhou, and Xipeng Qiu.
\newblock Speechgpt: Empowering large language models with intrinsic cross-modal conversational abilities, 2023.
\newblock URL \url{https://arxiv.org/abs/2305.11000}.

\bibitem[Microsoft et~al.(2025)Microsoft, :, Abouelenin, Ashfaq, Atkinson, Awadalla, Bach, Bao, Benhaim, Cai, Chaudhary, Chen, Chen, Chen, Chen, Chen, Chen, ling Chen, Dai, Dai, Fan, Gao, Gao, Garg, Goswami, Hao, Hendy, Hu, Jin, Khademi, Kim, Kim, Lee, Li, Li, Liang, Lin, Lin, Liu, Liu, Lopez, Luo, Madan, Mazalov, Mitra, Mousavi, Nguyen, Pan, Perez-Becker, Platin, Portet, Qiu, Ren, Ren, Roy, Shang, Shen, Singhal, Som, Song, Sych, Vaddamanu, Wang, Wang, Wang, Wu, Xu, Xu, Yang, Yang, Yu, Zabir, Zhang, Zhang, Zhang, and Zhou]{microsoft2025phi4minitechnicalreportcompact}
Microsoft, :, Abdelrahman Abouelenin, Atabak Ashfaq, Adam Atkinson, Hany Awadalla, Nguyen Bach, Jianmin Bao, Alon Benhaim, Martin Cai, Vishrav Chaudhary, Congcong Chen, Dong Chen, Dongdong Chen, Junkun Chen, Weizhu Chen, Yen-Chun Chen, Yi~ling Chen, Qi~Dai, Xiyang Dai, Ruchao Fan, Mei Gao, Min Gao, Amit Garg, Abhishek Goswami, Junheng Hao, Amr Hendy, Yuxuan Hu, Xin Jin, Mahmoud Khademi, Dongwoo Kim, Young~Jin Kim, Gina Lee, Jinyu Li, Yunsheng Li, Chen Liang, Xihui Lin, Zeqi Lin, Mengchen Liu, Yang Liu, Gilsinia Lopez, Chong Luo, Piyush Madan, Vadim Mazalov, Arindam Mitra, Ali Mousavi, Anh Nguyen, Jing Pan, Daniel Perez-Becker, Jacob Platin, Thomas Portet, Kai Qiu, Bo~Ren, Liliang Ren, Sambuddha Roy, Ning Shang, Yelong Shen, Saksham Singhal, Subhojit Som, Xia Song, Tetyana Sych, Praneetha Vaddamanu, Shuohang Wang, Yiming Wang, Zhenghao Wang, Haibin Wu, Haoran Xu, Weijian Xu, Yifan Yang, Ziyi Yang, Donghan Yu, Ishmam Zabir, Jianwen Zhang, Li~Lyna Zhang, Yunan Zhang, and Xiren Zhou.
\newblock Phi-4-mini technical report: Compact yet powerful multimodal language models via mixture-of-loras, 2025.
\newblock URL \url{https://arxiv.org/abs/2503.01743}.

\bibitem[KimiTeam et~al.(2025)KimiTeam, Ding, Ju, Leng, Liu, Liu, Shang, Shen, Song, Tan, Tang, Wang, Wei, Xin, Xu, Yu, Zhang, Zhou, Charles, Chen, Chen, Du, He, Hu, Lai, Li, Liu, Sun, Wang, Wang, Wu, Wu, Yang, Yang, Yang, Yang, Yin, Yuan, Zhang, and Zhou]{kimiteam2025kimiaudiotechnicalreport}
KimiTeam, Ding Ding, Zeqian Ju, Yichong Leng, Songxiang Liu, Tong Liu, Zeyu Shang, Kai Shen, Wei Song, Xu~Tan, Heyi Tang, Zhengtao Wang, Chu Wei, Yifei Xin, Xinran Xu, Jianwei Yu, Yutao Zhang, Xinyu Zhou, Y.~Charles, Jun Chen, Yanru Chen, Yulun Du, Weiran He, Zhenxing Hu, Guokun Lai, Qingcheng Li, Yangyang Liu, Weidong Sun, Jianzhou Wang, Yuzhi Wang, Yuefeng Wu, Yuxin Wu, Dongchao Yang, Hao Yang, Ying Yang, Zhilin Yang, Aoxiong Yin, Ruibin Yuan, Yutong Zhang, and Zaida Zhou.
\newblock Kimi-audio technical report, 2025.
\newblock URL \url{https://arxiv.org/abs/2504.18425}.

\bibitem[Fang et~al.(2025{\natexlab{a}})Fang, Guo, Zhou, Ma, Zhang, and Feng]{fang2024llamaomniseamlessspeechinteraction}
Qingkai Fang, Shoutao Guo, Yan Zhou, Zhengrui Ma, Shaolei Zhang, and Yang Feng.
\newblock {LL}a{MA}-omni: Seamless speech interaction with large language models.
\newblock In \emph{The Thirteenth International Conference on Learning Representations}, 2025{\natexlab{a}}.
\newblock URL \url{https://openreview.net/forum?id=PYmrUQmMEw}.

\bibitem[Xu et~al.(2025)Xu, Guo, He, Hu, He, Bai, Chen, Wang, Fan, Dang, Zhang, Wang, Chu, and Lin]{xu2025qwen25omnitechnicalreport}
Jin Xu, Zhifang Guo, Jinzheng He, Hangrui Hu, Ting He, Shuai Bai, Keqin Chen, Jialin Wang, Yang Fan, Kai Dang, Bin Zhang, Xiong Wang, Yunfei Chu, and Junyang Lin.
\newblock Qwen2.5-omni technical report, 2025.
\newblock URL \url{https://arxiv.org/abs/2503.20215}.

\bibitem[Fang et~al.(2025{\natexlab{b}})Fang, Zhou, Guo, Zhang, and Feng]{fang2025llamaomni2llmbasedrealtimespoken}
Qingkai Fang, Yan Zhou, Shoutao Guo, Shaolei Zhang, and Yang Feng.
\newblock Llama-omni2: Llm-based real-time spoken chatbot with autoregressive streaming speech synthesis, 2025{\natexlab{b}}.
\newblock URL \url{https://arxiv.org/abs/2505.02625}.

\bibitem[Zeng et~al.(2024)Zeng, Du, Liu, Wang, Jiang, Zhao, Dong, and Tang]{zeng2024glm4voiceintelligenthumanlikeendtoend}
Aohan Zeng, Zhengxiao Du, Mingdao Liu, Kedong Wang, Shengmin Jiang, Lei Zhao, Yuxiao Dong, and Jie Tang.
\newblock Glm-4-voice: Towards intelligent and human-like end-to-end spoken chatbot, 2024.
\newblock URL \url{https://arxiv.org/abs/2412.02612}.

\bibitem[Fang et~al.(2025{\natexlab{c}})Fang, Guo, Zhou, Ma, Zhang, and Feng]{fang2025llamaomniseamlessspeechinteraction}
Qingkai Fang, Shoutao Guo, Yan Zhou, Zhengrui Ma, Shaolei Zhang, and Yang Feng.
\newblock Llama-omni: Seamless speech interaction with large language models, 2025{\natexlab{c}}.
\newblock URL \url{https://arxiv.org/abs/2409.06666}.

\bibitem[Chung et~al.(2018)Chung, Nagrani, and Zisserman]{chung18b_interspeech}
Joon~Son Chung, Arsha Nagrani, and Andrew Zisserman.
\newblock Voxceleb2: Deep speaker recognition.
\newblock In \emph{Interspeech 2018}, pages 1086--1090, 2018.
\newblock \doi{10.21437/Interspeech.2018-1929}.

\bibitem[Wang et~al.(2021)Wang, Wu, Gu, and Pino]{wang21s_interspeech}
Changhan Wang, Anne Wu, Jiatao Gu, and Juan Pino.
\newblock Covost 2 and massively multilingual speech translation.
\newblock In \emph{Interspeech 2021}, pages 2247--2251, 2021.
\newblock \doi{10.21437/Interspeech.2021-2027}.

\bibitem[Gong et~al.(2023{\natexlab{a}})Gong, Liu, Luo, Karlinsky, and Glass]{2023jointaudiospeechunderstanding}
Yuan Gong, Alexander~H. Liu, Hongyin Luo, Leonid Karlinsky, and James~R. Glass.
\newblock Joint audio and speech understanding.
\newblock In \emph{{IEEE} Automatic Speech Recognition and Understanding Workshop, {ASRU} 2023, Taipei, Taiwan, December 16-20, 2023}, pages 1--8. {IEEE}, 2023{\natexlab{a}}.
\newblock \doi{10.1109/ASRU57964.2023.10389742}.
\newblock URL \url{https://doi.org/10.1109/ASRU57964.2023.10389742}.

\bibitem[Radford et~al.(2022)Radford, Kim, Xu, Brockman, McLeavey, and Sutskever]{radford2022robustspeechrecognitionlargescale}
Alec Radford, Jong~Wook Kim, Tao Xu, Greg Brockman, Christine McLeavey, and Ilya Sutskever.
\newblock Robust speech recognition via large-scale weak supervision, 2022.
\newblock URL \url{https://arxiv.org/abs/2212.04356}.

\bibitem[Graves et~al.(2006)Graves, Fern{\'a}ndez, Gomez, and Schmidhuber]{graves2006connectionist}
Alex Graves, Santiago Fern{\'a}ndez, Faustino Gomez, and J{\"u}rgen Schmidhuber.
\newblock Connectionist temporal classification: labelling unsegmented sequence data with recurrent neural networks.
\newblock In \emph{Proceedings of the 23rd international conference on Machine learning}, pages 369--376, 2006.

\bibitem[Bai et~al.(2023)Bai, Bai, Chu, Cui, Dang, Deng, Fan, Ge, Han, Huang, Hui, Ji, Li, Lin, Lin, Liu, Liu, Lu, Lu, Ma, Men, Ren, Ren, Tan, Tan, Tu, Wang, Wang, Wang, Wu, Xu, Xu, Yang, Yang, Yang, Yang, Yao, Yu, Yuan, Yuan, Zhang, Zhang, Zhang, Zhang, Zhou, Zhou, Zhou, and Zhu]{bai2023qwentechnicalreport}
Jinze Bai, Shuai Bai, Yunfei Chu, Zeyu Cui, Kai Dang, Xiaodong Deng, Yang Fan, Wenbin Ge, Yu~Han, Fei Huang, Binyuan Hui, Luo Ji, Mei Li, Junyang Lin, Runji Lin, Dayiheng Liu, Gao Liu, Chengqiang Lu, Keming Lu, Jianxin Ma, Rui Men, Xingzhang Ren, Xuancheng Ren, Chuanqi Tan, Sinan Tan, Jianhong Tu, Peng Wang, Shijie Wang, Wei Wang, Shengguang Wu, Benfeng Xu, Jin Xu, An~Yang, Hao Yang, Jian Yang, Shusheng Yang, Yang Yao, Bowen Yu, Hongyi Yuan, Zheng Yuan, Jianwei Zhang, Xingxuan Zhang, Yichang Zhang, Zhenru Zhang, Chang Zhou, Jingren Zhou, Xiaohuan Zhou, and Tianhang Zhu.
\newblock Qwen technical report, 2023.
\newblock URL \url{https://arxiv.org/abs/2309.16609}.

\bibitem[Rafailov et~al.(2023)Rafailov, Sharma, Mitchell, Manning, Ermon, and Finn]{rafailov2023direct}
Rafael Rafailov, Archit Sharma, Eric Mitchell, Christopher~D Manning, Stefano Ermon, and Chelsea Finn.
\newblock Direct preference optimization: Your language model is secretly a reward model.
\newblock In \emph{Thirty-seventh Conference on Neural Information Processing Systems}, 2023.
\newblock URL \url{https://openreview.net/forum?id=HPuSIXJaa9}.

\bibitem[Guo et~al.(2023)Guo, Zhang, and Feng]{guo2023simultaneousmachinetranslationtailored}
Shoutao Guo, Shaolei Zhang, and Yang Feng.
\newblock Simultaneous machine translation with tailored reference, 2023.
\newblock URL \url{https://arxiv.org/abs/2310.13588}.

\bibitem[Ren et~al.(2020)Ren, Liu, Tan, Zhang, Qin, Zhao, and Liu]{ren-etal-2020-simulspeech}
Yi~Ren, Jinglin Liu, Xu~Tan, Chen Zhang, Tao Qin, Zhou Zhao, and Tie-Yan Liu.
\newblock {S}imul{S}peech: End-to-end simultaneous speech to text translation.
\newblock In Dan Jurafsky, Joyce Chai, Natalie Schluter, and Joel Tetreault, editors, \emph{Proceedings of the 58th Annual Meeting of the Association for Computational Linguistics}, pages 3787--3796, Online, July 2020. Association for Computational Linguistics.
\newblock \doi{10.18653/v1/2020.acl-main.350}.
\newblock URL \url{https://aclanthology.org/2020.acl-main.350/}.

\bibitem[Zhang and Feng(2023)]{zhang-feng-2023-end}
Shaolei Zhang and Yang Feng.
\newblock End-to-end simultaneous speech translation with differentiable segmentation.
\newblock In Anna Rogers, Jordan Boyd-Graber, and Naoaki Okazaki, editors, \emph{Findings of the Association for Computational Linguistics: ACL 2023}, pages 7659--7680, Toronto, Canada, July 2023. Association for Computational Linguistics.
\newblock \doi{10.18653/v1/2023.findings-acl.485}.
\newblock URL \url{https://aclanthology.org/2023.findings-acl.485/}.

\bibitem[Lin et~al.(2024)Lin, Fu, Zhang, Liu, Zhang, Sun, Li, and Chen]{lin2024speechprunecontextawaretokenpruning}
Yueqian Lin, Yuzhe Fu, Jingyang Zhang, Yudong Liu, Jianyi Zhang, Jingwei Sun, Hai~"Helen" Li, and Yiran Chen.
\newblock Speechprune: Context-aware token pruning for speech information retrieval, 2024.
\newblock URL \url{https://arxiv.org/abs/2412.12009}.

\bibitem[Guo et~al.(2024)Guo, Zhang, and Feng]{Guo_2024}
Shoutao Guo, Shaolei Zhang, and Yang Feng.
\newblock Glancing future for simultaneous machine translation.
\newblock In \emph{ICASSP 2024 - 2024 IEEE International Conference on Acoustics, Speech and Signal Processing (ICASSP)}, page 11386–11390. IEEE, April 2024.
\newblock \doi{10.1109/icassp48485.2024.10446517}.
\newblock URL \url{http://dx.doi.org/10.1109/ICASSP48485.2024.10446517}.

\bibitem[Bai et~al.(2024)Bai, Lv, Zhang, Lyu, Tang, Huang, Du, Liu, Zeng, Hou, Dong, Tang, and Li]{bai-etal-2024-longbench}
Yushi Bai, Xin Lv, Jiajie Zhang, Hongchang Lyu, Jiankai Tang, Zhidian Huang, Zhengxiao Du, Xiao Liu, Aohan Zeng, Lei Hou, Yuxiao Dong, Jie Tang, and Juanzi Li.
\newblock {L}ong{B}ench: A bilingual, multitask benchmark for long context understanding.
\newblock In Lun-Wei Ku, Andre Martins, and Vivek Srikumar, editors, \emph{Proceedings of the 62nd Annual Meeting of the Association for Computational Linguistics (Volume 1: Long Papers)}, pages 3119--3137, Bangkok, Thailand, August 2024. Association for Computational Linguistics.
\newblock \doi{10.18653/v1/2024.acl-long.172}.
\newblock URL \url{https://aclanthology.org/2024.acl-long.172/}.

\bibitem[OpenAI(2024)]{gpt-4o}
OpenAI.
\newblock Hello gpt-4o, 2024.
\newblock URL \url{https://openai.com/index/hello-gpt-4o/}.

\bibitem[Panayotov et~al.(2015)Panayotov, Chen, Povey, and Khudanpur]{7178964}
Vassil Panayotov, Guoguo Chen, Daniel Povey, and Sanjeev Khudanpur.
\newblock Librispeech: An asr corpus based on public domain audio books.
\newblock In \emph{2015 IEEE International Conference on Acoustics, Speech and Signal Processing (ICASSP)}, pages 5206--5210, 2015.
\newblock \doi{10.1109/ICASSP.2015.7178964}.

\bibitem[Pratap et~al.(2020)Pratap, Xu, Sriram, Synnaeve, and Collobert]{Pratap_2020}
Vineel Pratap, Qiantong Xu, Anuroop Sriram, Gabriel Synnaeve, and Ronan Collobert.
\newblock Mls: A large-scale multilingual dataset for speech research.
\newblock In \emph{Interspeech 2020}. ISCA, October 2020.
\newblock \doi{10.21437/interspeech.2020-2826}.
\newblock URL \url{http://dx.doi.org/10.21437/Interspeech.2020-2826}.

\bibitem[Gong et~al.(2023{\natexlab{b}})Gong, Liu, Luo, Karlinsky, and Glass]{10389742}
Yuan Gong, Alexander~H. Liu, Hongyin Luo, Leonid Karlinsky, and James Glass.
\newblock Joint audio and speech understanding.
\newblock In \emph{2023 IEEE Automatic Speech Recognition and Understanding Workshop (ASRU)}, pages 1--8, 2023{\natexlab{b}}.
\newblock \doi{10.1109/ASRU57964.2023.10389742}.

\bibitem[Zhao et~al.(2024)Zhao, Jiang, Liu, Wang, and Wang]{zhao2024librisqanoveldatasetframework}
Zihan Zhao, Yiyang Jiang, Heyang Liu, Yanfeng Wang, and Yu~Wang.
\newblock Librisqa: A novel dataset and framework for spoken question answering with large language models, 2024.
\newblock URL \url{https://arxiv.org/abs/2308.10390}.

\bibitem[Ardila et~al.(2020)Ardila, Branson, Davis, Kohler, Meyer, Henretty, Morais, Saunders, Tyers, and Weber]{ardila-etal-2020-common}
Rosana Ardila, Megan Branson, Kelly Davis, Michael Kohler, Josh Meyer, Michael Henretty, Reuben Morais, Lindsay Saunders, Francis Tyers, and Gregor Weber.
\newblock Common voice: A massively-multilingual speech corpus.
\newblock In Nicoletta Calzolari, Fr{\'e}d{\'e}ric B{\'e}chet, Philippe Blache, Khalid Choukri, Christopher Cieri, Thierry Declerck, Sara Goggi, Hitoshi Isahara, Bente Maegaard, Joseph Mariani, H{\'e}l{\`e}ne Mazo, Asuncion Moreno, Jan Odijk, and Stelios Piperidis, editors, \emph{Proceedings of the Twelfth Language Resources and Evaluation Conference}, pages 4218--4222, Marseille, France, May 2020. European Language Resources Association.
\newblock ISBN 979-10-95546-34-4.
\newblock URL \url{https://aclanthology.org/2020.lrec-1.520/}.

\bibitem[Yang et~al.(2024)Yang, Xu, Liu, Chu, Jiang, Zhou, Leng, Lv, Zhao, Zhou, and Zhou]{yang-etal-2024-air}
Qian Yang, Jin Xu, Wenrui Liu, Yunfei Chu, Ziyue Jiang, Xiaohuan Zhou, Yichong Leng, Yuanjun Lv, Zhou Zhao, Chang Zhou, and Jingren Zhou.
\newblock {AIR}-bench: Benchmarking large audio-language models via generative comprehension.
\newblock In Lun-Wei Ku, Andre Martins, and Vivek Srikumar, editors, \emph{Proceedings of the 62nd Annual Meeting of the Association for Computational Linguistics (Volume 1: Long Papers)}, pages 1979--1998, Bangkok, Thailand, August 2024. Association for Computational Linguistics.
\newblock \doi{10.18653/v1/2024.acl-long.109}.
\newblock URL \url{https://aclanthology.org/2024.acl-long.109/}.

\bibitem[Poria et~al.(2019)Poria, Hazarika, Majumder, Naik, Cambria, and Mihalcea]{poria-etal-2019-meld}
Soujanya Poria, Devamanyu Hazarika, Navonil Majumder, Gautam Naik, Erik Cambria, and Rada Mihalcea.
\newblock {MELD}: A multimodal multi-party dataset for emotion recognition in conversations.
\newblock In Anna Korhonen, David Traum, and Llu{\'i}s M{\`a}rquez, editors, \emph{Proceedings of the 57th Annual Meeting of the Association for Computational Linguistics}, pages 527--536, Florence, Italy, July 2019. Association for Computational Linguistics.
\newblock \doi{10.18653/v1/P19-1050}.
\newblock URL \url{https://aclanthology.org/P19-1050/}.

\bibitem[Lu et~al.(2024)Lu, Song, Yang, and Watanabe]{lu2024fastadaspmultitaskadaptedefficientinference}
Yichen Lu, Jiaqi Song, Chao-Han~Huck Yang, and Shinji Watanabe.
\newblock Fastadasp: Multitask-adapted efficient inference for large speech language model, 2024.
\newblock URL \url{https://arxiv.org/abs/2410.03007}.

\bibitem[Chen et~al.(2021)Chen, Chai, Wang, Du, Zhang, Weng, Su, Povey, Trmal, Zhang, Jin, Khudanpur, Watanabe, Zhao, Zou, Li, Yao, Wang, Wang, You, and Yan]{chen2021gigaspeechevolvingmultidomainasr}
Guoguo Chen, Shuzhou Chai, Guanbo Wang, Jiayu Du, Wei-Qiang Zhang, Chao Weng, Dan Su, Daniel Povey, Jan Trmal, Junbo Zhang, Mingjie Jin, Sanjeev Khudanpur, Shinji Watanabe, Shuaijiang Zhao, Wei Zou, Xiangang Li, Xuchen Yao, Yongqing Wang, Yujun Wang, Zhao You, and Zhiyong Yan.
\newblock Gigaspeech: An evolving, multi-domain asr corpus with 10,000 hours of transcribed audio, 2021.
\newblock URL \url{https://arxiv.org/abs/2106.06909}.

\bibitem[Lin et~al.(2025)Lin, Fu, Zhang, Liu, Zhang, Sun, Li, and Chen]{lin2025speechprunecontextawaretokenpruning}
Yueqian Lin, Yuzhe Fu, Jingyang Zhang, Yudong Liu, Jianyi Zhang, Jingwei Sun, Hai~"Helen" Li, and Yiran Chen.
\newblock Speechprune: Context-aware token pruning for speech information retrieval, 2025.
\newblock URL \url{https://arxiv.org/abs/2412.12009}.

\bibitem[Su et~al.(2023)Su, Lu, Pan, Murtadha, Wen, and Liu]{su2023roformerenhancedtransformerrotary}
Jianlin Su, Yu~Lu, Shengfeng Pan, Ahmed Murtadha, Bo~Wen, and Yunfeng Liu.
\newblock Roformer: Enhanced transformer with rotary position embedding, 2023.
\newblock URL \url{https://arxiv.org/abs/2104.09864}.

\bibitem[bloc97(2023)]{bloc97_2023}
bloc97.
\newblock Ntk-aware scaled rope allows llama models to have extended (8k+) context size without any fine-tuning and minimal perplexity degradation, 2023.

\bibitem[Hu et~al.(2021)Hu, Shen, Wallis, Allen-Zhu, Li, Wang, Wang, and Chen]{hu2021lora}
Edward~J. Hu, Yelong Shen, Phillip Wallis, Zeyuan Allen-Zhu, Yuanzhi Li, Shean Wang, Lu~Wang, and Weizhu Chen.
\newblock Lora: Low-rank adaptation of large language models, 2021.

\bibitem[Ding et~al.(2022)Ding, Zhang, Han, and Ding]{Ding_2022_CVPR}
Xiaohan Ding, Xiangyu Zhang, Jungong Han, and Guiguang Ding.
\newblock Scaling up your kernels to 31x31: Revisiting large kernel design in cnns.
\newblock In \emph{Proceedings of the IEEE/CVF Conference on Computer Vision and Pattern Recognition (CVPR)}, pages 11963--11975, June 2022.

\bibitem[Prabhavalkar et~al.(2024)Prabhavalkar, Hori, Sainath, Schlüter, and Watanabe]{10301513}
Rohit Prabhavalkar, Takaaki Hori, Tara~N. Sainath, Ralf Schlüter, and Shinji Watanabe.
\newblock End-to-end speech recognition: A survey.
\newblock \emph{IEEE/ACM Transactions on Audio, Speech, and Language Processing}, 32:\penalty0 325--351, 2024.
\newblock \doi{10.1109/TASLP.2023.3328283}.

\bibitem[Fu et~al.(2024)Fu, Lin, Long, Shen, Zhao, Zhang, Dong, Wang, Yin, Ma, Zheng, He, Ji, Wu, Shan, and Sun]{fu2024vitaopensourceinteractiveomni}
Chaoyou Fu, Haojia Lin, Zuwei Long, Yunhang Shen, Meng Zhao, Yifan Zhang, Shaoqi Dong, Xiong Wang, Di~Yin, Long Ma, Xiawu Zheng, Ran He, Rongrong Ji, Yunsheng Wu, Caifeng Shan, and Xing Sun.
\newblock Vita: Towards open-source interactive omni multimodal llm, 2024.
\newblock URL \url{https://arxiv.org/abs/2408.05211}.

\bibitem[Rubenstein et~al.(2023)Rubenstein, Asawaroengchai, Nguyen, Bapna, Borsos, de~Chaumont~Quitry, Chen, Badawy, Han, Kharitonov, Muckenhirn, Padfield, Qin, Rozenberg, Sainath, Schalkwyk, Sharifi, Ramanovich, Tagliasacchi, Tudor, Velimirović, Vincent, Yu, Wang, Zayats, Zeghidour, Zhang, Zhang, Zilka, and Frank]{rubenstein2023audiopalmlargelanguagemodel}
Paul~K. Rubenstein, Chulayuth Asawaroengchai, Duc~Dung Nguyen, Ankur Bapna, Zalán Borsos, Félix de~Chaumont~Quitry, Peter Chen, Dalia~El Badawy, Wei Han, Eugene Kharitonov, Hannah Muckenhirn, Dirk Padfield, James Qin, Danny Rozenberg, Tara Sainath, Johan Schalkwyk, Matt Sharifi, Michelle~Tadmor Ramanovich, Marco Tagliasacchi, Alexandru Tudor, Mihajlo Velimirović, Damien Vincent, Jiahui Yu, Yongqiang Wang, Vicky Zayats, Neil Zeghidour, Yu~Zhang, Zhishuai Zhang, Lukas Zilka, and Christian Frank.
\newblock Audiopalm: A large language model that can speak and listen, 2023.
\newblock URL \url{https://arxiv.org/abs/2306.12925}.

\bibitem[Défossez et~al.(2024)Défossez, Mazaré, Orsini, Royer, Pérez, Jégou, Grave, and Zeghidour]{défossez2024moshispeechtextfoundationmodel}
Alexandre Défossez, Laurent Mazaré, Manu Orsini, Amélie Royer, Patrick Pérez, Hervé Jégou, Edouard Grave, and Neil Zeghidour.
\newblock Moshi: a speech-text foundation model for real-time dialogue, 2024.
\newblock URL \url{https://arxiv.org/abs/2410.00037}.

\bibitem[Chen et~al.(2023)Chen, Wong, Chen, and Tian]{chen2023extendingcontextwindowlarge}
Shouyuan Chen, Sherman Wong, Liangjian Chen, and Yuandong Tian.
\newblock Extending context window of large language models via positional interpolation, 2023.
\newblock URL \url{https://arxiv.org/abs/2306.15595}.

\bibitem[Ratner et~al.(2023)Ratner, Levine, Belinkov, Ram, Magar, Abend, Karpas, Shashua, Leyton-Brown, and Shoham]{ratner-etal-2023-parallel}
Nir Ratner, Yoav Levine, Yonatan Belinkov, Ori Ram, Inbal Magar, Omri Abend, Ehud Karpas, Amnon Shashua, Kevin Leyton-Brown, and Yoav Shoham.
\newblock Parallel context windows for large language models.
\newblock In Anna Rogers, Jordan Boyd-Graber, and Naoaki Okazaki, editors, \emph{Proceedings of the 61st Annual Meeting of the Association for Computational Linguistics (Volume 1: Long Papers)}, pages 6383--6402, Toronto, Canada, July 2023. Association for Computational Linguistics.
\newblock \doi{10.18653/v1/2023.acl-long.352}.
\newblock URL \url{https://aclanthology.org/2023.acl-long.352/}.

\bibitem[Rozière et~al.(2024)Rozière, Gehring, Gloeckle, Sootla, Gat, Tan, Adi, Liu, Sauvestre, Remez, Rapin, Kozhevnikov, Evtimov, Bitton, Bhatt, Ferrer, Grattafiori, Xiong, Défossez, Copet, Azhar, Touvron, Martin, Usunier, Scialom, and Synnaeve]{rozière2024codellamaopenfoundation}
Baptiste Rozière, Jonas Gehring, Fabian Gloeckle, Sten Sootla, Itai Gat, Xiaoqing~Ellen Tan, Yossi Adi, Jingyu Liu, Romain Sauvestre, Tal Remez, Jérémy Rapin, Artyom Kozhevnikov, Ivan Evtimov, Joanna Bitton, Manish Bhatt, Cristian~Canton Ferrer, Aaron Grattafiori, Wenhan Xiong, Alexandre Défossez, Jade Copet, Faisal Azhar, Hugo Touvron, Louis Martin, Nicolas Usunier, Thomas Scialom, and Gabriel Synnaeve.
\newblock Code llama: Open foundation models for code, 2024.
\newblock URL \url{https://arxiv.org/abs/2308.12950}.

\bibitem[Yuan et~al.(2025)Yuan, Gao, Dai, Luo, Zhao, Zhang, Xie, Wei, Wang, Xiao, Wang, Ruan, Zhang, Liang, and Zeng]{yuan2025nativesparseattentionhardwarealigned}
Jingyang Yuan, Huazuo Gao, Damai Dai, Junyu Luo, Liang Zhao, Zhengyan Zhang, Zhenda Xie, Y.~X. Wei, Lean Wang, Zhiping Xiao, Yuqing Wang, Chong Ruan, Ming Zhang, Wenfeng Liang, and Wangding Zeng.
\newblock Native sparse attention: Hardware-aligned and natively trainable sparse attention, 2025.
\newblock URL \url{https://arxiv.org/abs/2502.11089}.

\bibitem[Song et~al.(2024)Song, Chai, Wang, Zhang, Zhou, Wu, Chi, Guo, Ye, Zhang, Lu, Hwang, and Wang]{song2024moviechatdensetokensparse}
Enxin Song, Wenhao Chai, Guanhong Wang, Yucheng Zhang, Haoyang Zhou, Feiyang Wu, Haozhe Chi, Xun Guo, Tian Ye, Yanting Zhang, Yan Lu, Jenq-Neng Hwang, and Gaoang Wang.
\newblock Moviechat: From dense token to sparse memory for long video understanding, 2024.
\newblock URL \url{https://arxiv.org/abs/2307.16449}.

\bibitem[Shang et~al.(2024)Shang, Cai, Xu, Lee, and Yan]{shang2024llavaprumergeadaptivetokenreduction}
Yuzhang Shang, Mu~Cai, Bingxin Xu, Yong~Jae Lee, and Yan Yan.
\newblock Llava-prumerge: Adaptive token reduction for efficient large multimodal models, 2024.
\newblock URL \url{https://arxiv.org/abs/2403.15388}.

\bibitem[Tsunoo et~al.(2024)Tsunoo, Futami, Kashiwagi, Arora, and Watanabe]{tsunoo2024decoderonlyarchitecturespeechrecognition}
Emiru Tsunoo, Hayato Futami, Yosuke Kashiwagi, Siddhant Arora, and Shinji Watanabe.
\newblock Decoder-only architecture for speech recognition with ctc prompts and text data augmentation, 2024.
\newblock URL \url{https://arxiv.org/abs/2309.08876}.

\bibitem[Gaido et~al.(2021)Gaido, Cettolo, Negri, and Turchi]{gaido2021ctcbasedcompressiondirectspeech}
Marco Gaido, Mauro Cettolo, Matteo Negri, and Marco Turchi.
\newblock Ctc-based compression for direct speech translation, 2021.
\newblock URL \url{https://arxiv.org/abs/2102.01578}.

\end{thebibliography}


\appendix


\section{Description of LongSpeech-Eval}
\label{longbench}
LongSpeech-Eval is a novel benchmark we propose for evaluating the long-speech understanding capabilities of Large Speech-Language Models (LSLMs). This benchmark presents a spoken Question-Answering (QA) task, challenging LSLMs to answer questions based on the extended speech inputs. The dataset comprises 164 samples, with an average speech duration of 132.77 seconds and a maximum duration reaching 1000 seconds.

The foundation for LongSpeech-Eval is the MultiField-En and NarrativeQA subsets from LongBench, an established long-context understanding benchmark. MultiField-En is a single-document QA dataset encompassing diverse domains, with questions and answers meticulously annotated by Ph.D. students. NarrativeQA consists of long stories along
with questions posed to test reading comprehension. Our methodology for creating LongSpeech-Eval involves a rigorous multi-step process.

We first employ Llama3.1-70B-Instruct\footnote{\url{https://huggingface.co/meta-llama/Llama-3.1-70B-Instruct}} to filter out samples containing numerous formulas or non-English characters, ensuring the dataset's suitability for speech synthesis and comprehension. GPT-4o \citep{gpt-4o} is utilized to summarize and polish the documents into more natural spoken forms, enhancing their suitability for speech synthesis. We then reapply Llama3.1-70B-Instruct to eliminate any samples where questions could not be adequately answered based on the spoken-form documents, ensuring the validity of the samples. Finally, we leverage the Text-to-Speech (TTS) model Orca\footnote{\url{https://github.com/Picovoice/orca}} to synthesize speech from the refined spoken-form documents.

The resulting dataset combines synthesized speech with corresponding questions and answers, forming a comprehensive spoken QA benchmark.

\section{Details of Dataset}
\label{dataset}
In this section, we provide a detailed description of the training and testing data.
\subsection{Training Dataset}
Our training method is divided into two stages. 

In the first stage, we train the CTC decoder using the CTC loss \citep{radford2022robustspeechrecognitionlargescale}. During this stage, only ASR data are used, including 960 hours of LibriSpeech \citep{7178964} data and 3k hours of data sampled from MLS dataset \citep{Pratap_2020}. 

In the second stage, we utilize our proposed dynamic compression training approach to train the LLM. For this stage, we use spoken QA datasets, which come from three datasets: OpenASQA \citep{10389742}, LibriSQA \citep{zhao2024librisqanoveldatasetframework}, and Common Voice \citep{ardila-etal-2020-common}. For \textbf{OpenASQA}, we select the Open-Ended Speech AQA subset, which contains 5.9k hours of speech data. The questions and answers in this dataset are generated by GPT-3.5-Turbo and cover aspects such as spoken text, speaker gender, age, style, and emotion. For \textbf{LibriSQA}, we use the complete training set, which contains 360 hours of training data. The questions and answers in this dataset are generated by ChatGPT, with the speech data sourced from the LibriSpeech train-clean-360 subset \citep{zhao2024librisqanoveldatasetframework}. For the \textbf{Common Voice} ASR dataset, we transform it into a spoken QA format to enhance our training set. First, we use ChatGPT to generate 200 diverse speech transcription instructions. For each ASR sample, we randomly select one instruction as the question and use the ground-truth transcription as the answer, resulting in 1.7k hours of training data.

\subsection{Evaluation Dataset}
For testing, we evaluate our method on short-speech spoken QA, long-speech spoken QA, and ASR tasks. The long-speech spoken QA task corresponds to the LongSpeech-Eval benchmark, which is introduced in Appendix \ref{longbench}. 

For short-speech spoken QA, we utilize three test sets: the speech\_QA\_iemocap (AIR-Bench) \citep{yang-etal-2024-air}, the LibriSQA test set \citep{zhao2024librisqanoveldatasetframework}, and the LibriTTS test subset from OpenASQA \citep{10389742}. The speech\_QA\_iemocap dataset comes from the AIR-Bench benchmark and contains 200 samples. The LibriSQA test set includes 2620 samples. For the LibriTTS test subset, we select samples corresponding to the LibriTTS test-clean set from OpenASQA, keeping only the 417 samples with a speech duration longer than 15 seconds as our test set. All test sets are under 30s in duration.

For spoken dialogue understanding, we evaluate the inference efficiency of our method using speech\_dialogue\_QA\_fisher subset \citep{yang-etal-2024-air} from AIR-Bench. This subset contains 200 samples. For this task, our method is directly applied to vanilla Qwen2-Audio, which has only undergone the first training phase of our method. This setup allows us to assess the effectiveness of our approach without requiring the training of LSLMs.

For the ASR task, we use the LibriSpeech \citep{7178964} test-clean, test-other, and GigaSpeech \citep{chen2021gigaspeechevolvingmultidomainasr} test set as our evaluation datasets. For convenience in evaluation, we convert these datasets into the spoken QA format, where the instruction for each sample is: ``\textbf{Transcribe the speech to text without explanation:} ''.

For emotion recognition task, We leverage the MELD dataset \citep{poria-etal-2019-meld} to benchmark our method against other efficiency method \citep{lu2024fastadaspmultitaskadaptedefficientinference} under diverse efficiency scenarios. FastAdaSP lowers the inference costs of Qwen2-Audio through the layer-wise dynamic reduction of speech representations within the LLM's architecture. We compare our method with FastAdaSP to demonstrate the advantage of our method in retaining information. Since we could not find the specific prompt in FastAdaSP \citep{lu2024fastadaspmultitaskadaptedefficientinference}, we utilize the following prompt: "\textbf{Given the Choices: [Anger, Disgust, Fear, Joy, Neutral, Sadness, Surprise]. What is the emotion in the audio?}"

\begin{table*}[t]
\centering
\caption{Settings of FastLongSpeech.}
\begin{tabular}{c|c|c|c}
\toprule
\multicolumn{3}{c|}{\textbf{Hyperparameters}} & \textbf{Settings} \\
\midrule

\multirow{5}{*}{CTC Decoder}  & \multirow{2}{*}{Model}  & hidden\_dim & 4096 \\
                         &    & output\_dim & 10000 \\
\cmidrule{2-4}
 & \multirow{3}{*}{Training Details}  & per\_device\_batch\_size & 16  \\
                     &     & learning\_rate & 2e-5 \\
                     &     & lr\_scheduler & cosine \\
                    \cmidrule{1-4}
\multirow{9}{*}{LSLM} &  Base\_model & Base\_model & \texttt{Qwen2-Audio-7B-Instruct} \\
                    \cmidrule{2-4}
                    &  \multirow{4}{*}{LoRA}  & lora\_r  & 128 \\
                     &     & lora\_alpha & 256   \\
                     &     & lora\_dropout & 0.05 \\
                     &     & lora\_target\_modules & q\_proj, k\_proj, v\_proj, o\_proj \\
                     \cmidrule{2-4}
                    &  \multirow{3}{*}{Training Details} & per\_device\_batch\_size & 16  \\
                     &     & learning\_rate & 2e-4 \\
                     &     & lr\_scheduler & cosine \\

\bottomrule

\end{tabular}
\label{hyperparameter}
\end{table*}

\section{Experimental Details}
\label{experiment}
In this section, we introduce the NTK-RoPE method in greater detail and outline the system configuration of FastLongSpeech. Our FastLongSpeech primarily leverages Qwen2-Audio. Additionally, we apply our approach to a vanilla Qwen2.5-Omni \citep{xu2025qwen25omnitechnicalreport} model that has only undergone the first training phase. This serves to validate that our method can achieve competitive performance without altering the inherent capabilities of the model, while also demonstrating its generalizability.

NTK-RoPE extends the speech window of Qwen2-Audio to match the context length of its LLM by adjusting the Rotary Position Embedding (RoPE). \textbf{However, some samples in our LongSpeech-Eval may still exceed this extended context length. To handle these special cases, we apply our iterative fusion strategy to reduce the sequence of speech representations to fit within the prescribed context length}.

We then delineate the configuration of FastLongSpeech. The training process is in two stages.
In the first stage, we utilize ASR data to train the CTC Decoder, which is a feed-forward network with one hidden layer. We use the SentencePiece\footnote{\url{https://github.com/google/sentencepiece}} toolkit to construct the vocabulary for the training of the CTC decoder. This vocabulary is extracted from the ASR dataset.
The second stage focuses on training the LLM within the LSLM using Spoken QA data. Both training stages leverage DeepSpeed\footnote{\url{https://github.com/deepspeedai/DeepSpeed}} ZeRO-2 for optimization. Table \ref{hyperparameter} provides additional training and configuration details.

\section{Evaluation Template}
\label{evaluation_template}
\begin{figure*}[t]
    \centering
    \includegraphics[width=6in]{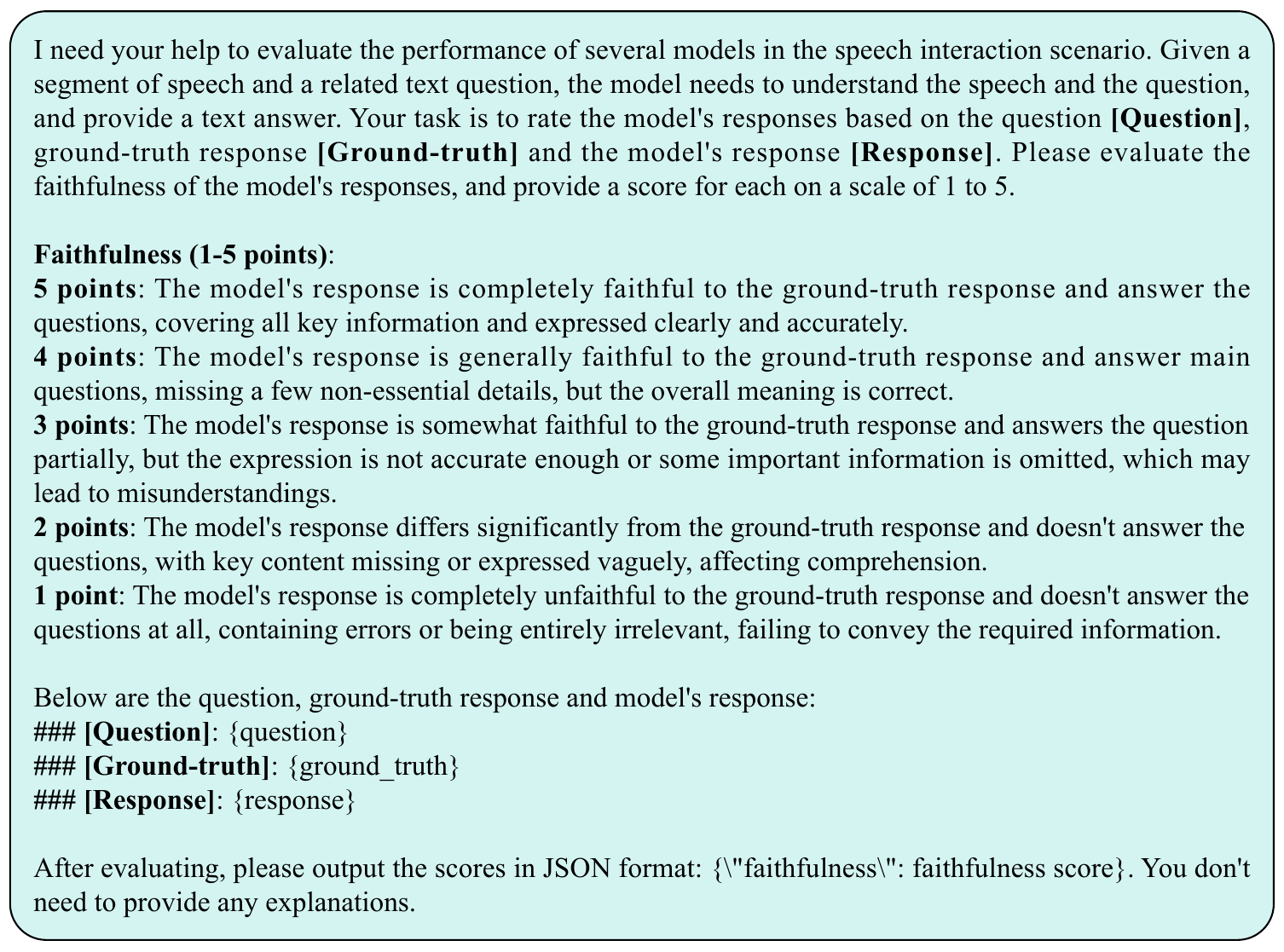}
    \caption{The prompt template for the LLM to evaluate the response of LSLMs.}
    \label{Instruct}
\end{figure*}

In this section, we present the prompt template used for evaluating LSLMs. As shown in Figure \ref{Instruct}, the template will be employed by the LLM to score the responses generated by the LSLMs. This scoring template is used to evaluate long-speech and short-speech spoken QA tasks.

\section{Applicability of Our Method to Vanilla LSLMs}
\label{vanilla_methods}
In our FastLongSpeech framework, we extend LSLMs for long-speech processing by adopting an iterative fusion strategy and a dynamic compression training approach. 
As highlighted in subsection \ref{infer_efficiency}, our method not only excels in long-speech tasks but also achieves a good balance between performance and efficiency in short-speech scenarios. Therefore, this prompts us to investigate whether vanilla LSLMs can benefit from our method to effectively balance computational efficiency and generation quality, thus meeting diverse requirements across various speech processing applications. 
To this end, we apply the iterative fusion strategy directly to the vanilla Qwen2-Audio and vanilla Qwen2.5-Omni model.

\begin{wraptable}{r}{6.0cm}
\vspace{-0.2in} 
\caption{The experiment results on the speech\_dialogue\_QA\_fisher subset, where ``Baseline'' denotes vanilla Qwen2-Audio and ``Ours'' denotes applying iterate fusion strategy to vanilla Qwen2-Audio.}
\small
\centering
\label{application}
\begin{tabular}{c c c} \toprule[1.2pt]
\textbf{Method} & \textbf{Score} ($\uparrow$) & \textbf{TFLOPs} ($\downarrow$) \\ 

\cmidrule(lr){1-1} \cmidrule(lr){2-3}

Baseline & 3.95 & 11.76 \\
\cmidrule(lr){1-1} \cmidrule(lr){2-3}
Ours ($L$=400) & 4.13 & 8.25 \\
Ours ($L$=200) & 3.92 & 5.46 \\
Ours ($L$=100) & 3.62 & 4.06 \\
Ours ($L$=50) & 3.16 & 3.35 \\

 \bottomrule[1pt]
\end{tabular}
\vspace{-0.1in} 
\end{wraptable}

To demonstrate the effectiveness and robustness of our method, we first extend our experiments to spoken dialogue understanding task. For this task, we conduct experiments on vanilla Qwen2-Audio using speech\_dialogue\_QA\_fisher of AIR-Bench \citep{yang-etal-2024-air}. As shown in Table \ref{application}, our method effectively balances performance and inference efficiency. Notably, at lower compression ratios ($L$ = 200), our approach demonstrate comparable performance to the vanilla Qwen2-Audio model with a 50\% reduction in computational costs. Moreover, even at a higher compression ratio of 15x ($L$=50), our method still maintains robust performance. These findings underscore the efficacy and versatility of our iterative fusion strategy.

\begin{wraptable}{r}{6.0cm}
\vspace{-0.2in} 
\caption{The experiment results on the speech\_QA\_iemocap subset.}
\small
\centering
\label{qwen2_5omni}
\begin{tabular}{c c c} \toprule[1.2pt]
$L$ & \textbf{Qwen2-Audio} & \textbf{Qwen2.5-Omni} \\

\cmidrule(lr){1-1} \cmidrule(lr){2-3}

750 & 3.68 & 3.82  \\

400 & 3.69 & 3.82  \\

200 & 3.67 & 3.75 \\ 

 \bottomrule[1pt]
\end{tabular}
\vspace{-0.1in} 
\end{wraptable}

We further extend our approach to vanilla Qwen2.5-Omni \citep{xu2025qwen25omnitechnicalreport}, a model exhibiting superior capabilities compared to Qwen2-Audio. Specifically, we benchmark the performance of Qwen2.5-Omni against Qwen2-Audio on the speech\_QA\_iemocap subset of AIR-Bench. The results in Table \ref{qwen2_5omni} indicate that Qwen2.5-Omni, owing to its stronger speech capabilities, demonstrates superior performance across various compression ratios. This demonstrates that our method achieves superior performance on more capable LSLMs, highlighting its generalizability.

\begin{wraptable}{r}{5.8cm}
\vspace{-0.2in} 
\caption{The experiments on the MELD dataset, where the results are reported in the configuration with a 50\% reduction in inference cost. The performance is measured with accuracy metric.}
\small
\centering
\label{meld}
\begin{tabular}{c c c} \toprule[1.2pt]
\textbf{Method} & \textbf{Accuracy (\%)} ($\uparrow$) \\ 

\cmidrule(lr){1-1} \cmidrule(lr){2-2}

FastAdaSP & 52.14 \\

\textbf{FastLongSpeech} & \textbf{52.95} \\

 \bottomrule[1pt]
\end{tabular}
\vspace{-0.2in} 
\end{wraptable}

\section{Applicability of Our Method to Other Tasks}
\label{meld_section}
Additionally, we extend our experimental evaluation to the emotion recognition task and employ MELD dataset  \citep{poria-etal-2019-meld}. For this task, we benchmark our method against FastAdaSP \citep{lu2024fastadaspmultitaskadaptedefficientinference}, an approach designed for enhancing inference efficiency of Qwen2-Audio. We adopt the identical experimental setup as in the FastAdaSP, comparing performance under the same inference reduction settings. As depicted in Table \ref{meld}, our method not only achieved superior performance but also reduced inference cost by 50\% compared to FastAdaSP-Sparse. This underscores the effectiveness of our approach in preserving crucial information. Furthermore, our method is complementary to the method \citep{lu2024fastadaspmultitaskadaptedefficientinference} and holds potential for further improving inference efficiency through integration, a prospect we leave for future investigation.

\begin{table}
\caption{The experiments on the SPIRAL-H dataset. The performance is measured with accuracy metric.}
\small
\centering
\label{speechprune}
\begin{tabular}{c c c} \toprule[1.2pt]
\textbf{Method} & \textbf{Prune Rate (\%)}($\uparrow$)  & \textbf{Accuracy (\%)} ($\uparrow$) \\ 

\cmidrule(lr){1-1} \cmidrule(lr){2-3}

SpeechPrune & 60.00 & 63.77 \\

\textbf{FastLongSpeech ($L=750$)} & \textbf{65.88} & \textbf{76.81} \\

 \bottomrule[1pt]
\end{tabular}
\end{table}
Beyond emotion recognition, we also conduct additional experiments on the SPIRAL-H\footnote{\url{https://github.com/linyueqian/SPIRAL_Dataset}} dataset, a benchmark designed for long speech information retrieval. On this dataset, we follow the experimental setup \citep{lin2024speechprunecontextawaretokenpruning} and compare model performance under similar speech embedding pruning rates. As shown in the table \ref{speechprune}, our method achieves better performance with fewer speech embeddings, demonstrating its superiority and efficiency in modeling long speech inputs.

\begin{table}
\caption{The performance of FastLongSpeech with varying context lengths.}
\small
\centering
\label{vary_context_len}
\begin{tabular}{c c c c c c c} \toprule[1.2pt]
$\mathbf{L}$ & 200 & 400 & 750 & 1200 & 4000 \\
\cmidrule(lr){1-1} \cmidrule(lr){2-6} 

Score ($\uparrow$) & 2.47 & 2.90 & 3.55 & 3.66 & 3.59 \\

 \bottomrule[1pt]
\end{tabular}
\end{table}
\section{Extending Maximum Context Length}
We also explore the performance of FastLongSpeech on the LongSpeech-Eval dataset with varying context lengths $\mathbf{L}$. The results are shown in Table \ref{vary_context_len}. When $L$ is less than 750, the model exhibits increasing performance with longer context, as it is trained under dynamically varying compression ratios in this range. When $L$ equals 1200, although the model is not explicitly trained for this length, it still achieves strong performance, indicating good generalization beyond the training regime. When increases to 4000, performance slightly declines. This is expected, as the model is not exposed to such long contexts during training, despite having a larger speech context window. We think our FastLongSpeech can achieve better performance with longer effective context length as the long-speech training data becomes more available.

\end{document}